\documentclass[10pt,twocolumn,letterpaper]{article}

\usepackage{iccv}
\usepackage{times}
\usepackage{epsfig}
\usepackage{graphicx}
\usepackage{amsmath}
\usepackage{amssymb}
\usepackage{algorithmicx}
\usepackage{algorithm}
\usepackage{algpseudocode}
\usepackage{caption}
\usepackage{epstopdf}
\usepackage{subcaption}
\usepackage{booktabs}

\usepackage[pagebackref=true,breaklinks=true,letterpaper=true,colorlinks,bookmarks=false]{hyperref}

\iccvfinalcopy 

\ificcvfinal\fi
\begin{document}

\title{Multi-task Dictionary Learning based Convolutional Neural Network for Computer aided Diagnosis with Longitudinal Images}

\author{Jie Zhang$^1$, Qingyang Li$^1$, Richard J. Caselli$^2$, Jieping Ye$^3$, Yalin Wang$^1$\\
$^1$School of Computing, Informatics, and Decision Systems Engineering, Arizona State University, \\Tempe, AZ, $^2$Department of Neurology, Mayo Clinic, Scottsdale, AZ, \\$^3$Department of Electrical Engineering and Computer Science, University of Michigan, Ann Arbor, MI\\
{\tt\small \{JieZhang.Joena, Qingyang.Li, Yalin.Wang\}@asu.edu,}
{\tt\small jpye@umich.edu,}
{\tt\small caselli.richard@mayo.edu}
}
\maketitle

\begin{abstract}
Algorithmic image-based diagnosis and prognosis of neurodegenerative diseases on longitudinal data has drawn great interest from computer vision researchers. The current state-of-the-art models for many image classification tasks are based on the Convolutional Neural Networks (CNN). However, a key challenge in applying CNN to biological problems is that the available labeled training samples are very limited. Another issue for CNN to be applied in computer aided diagnosis applications is that to achieve better diagnosis and prognosis accuracy, one usually has to deal with the longitudinal dataset,  i.e., the dataset of images scanned at different time points. Here we argue that an enhanced CNN model with transfer learning for the joint analysis of tasks from multiple time points or regions of interests may have a potential to improve the accuracy of computer aided diagnosis. To reach this goal, we innovate a CNN based deep learning multi-task dictionary learning framework to address the above challenges. Firstly, we pre-train CNN on the ImageNet dataset and transfer the knowledge from the pre-trained model to the medical imaging progression representation, generating the features for different tasks. Then, we propose a novel unsupervised learning method, termed Multi-task Stochastic Coordinate Coding (MSCC), for learning different tasks by using shared and individual dictionaries and generating the sparse features required to predict the future cognitive clinical scores. We apply our new model in a publicly available neuroimaging cohort to predict clinical measures with two different feature sets and compare them with seven other state-of-the-art methods. The experimental results show our proposed method achieved superior results.
\end{abstract}

\vspace{-1.5em}
\section{Introduction}
Deep learning models \cite{sharif2014cnn, zeiler2014visualizing, zhang2015deep} are capable of learning the hierarchical structure of features extracted from real-world images. Convolutional Neural Networks (CNNs) are a class of multi-layer, fully trainable models that are able to capture highly nonlinear mappings between inputs and outputs~\cite{lecun1998gradient}. Recently, CNNs have been successfully applied to a variety of applications, including image classification~\cite{krizhevsky2012imagenet}, segmentation~\cite{turaga2010convolutional}, and biological problems~\cite{hazlett2017early}. Feature learning with deep learning model typically requires a large amount of training data. Thus, feature learning for domains with scarce data is not feasible. However, a key challenge in applying CNNs to biological problems is that the available labeled training samples are very limited. Transfer learning \cite{blitzer2006domain, pan2010survey, zhang2011deep} is one of the approaches to address this problem and help feature learning in the data-scarce target domain by transferring knowledge from data-rich source domain. In this study, we aim to explore whether the nice transfer learning property of CNN can be help apply CNN to general biological image researches.

\begin{figure}[t]
\centering
\includegraphics[height=5cm]{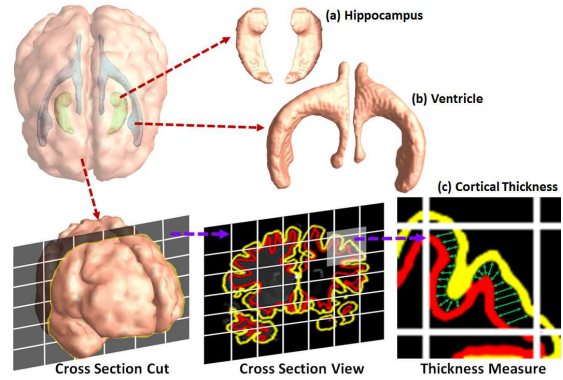}
\caption{This figure shows three promising anatomical features of the brain structural MR images used for clinical diagnosis of Alzheimer's Disease.}
\vspace{-1.5em}
\label{fig:1}
\end{figure}

\begin{figure*}[t]
\vspace{-1em}
\centering
\includegraphics[height=8cm]{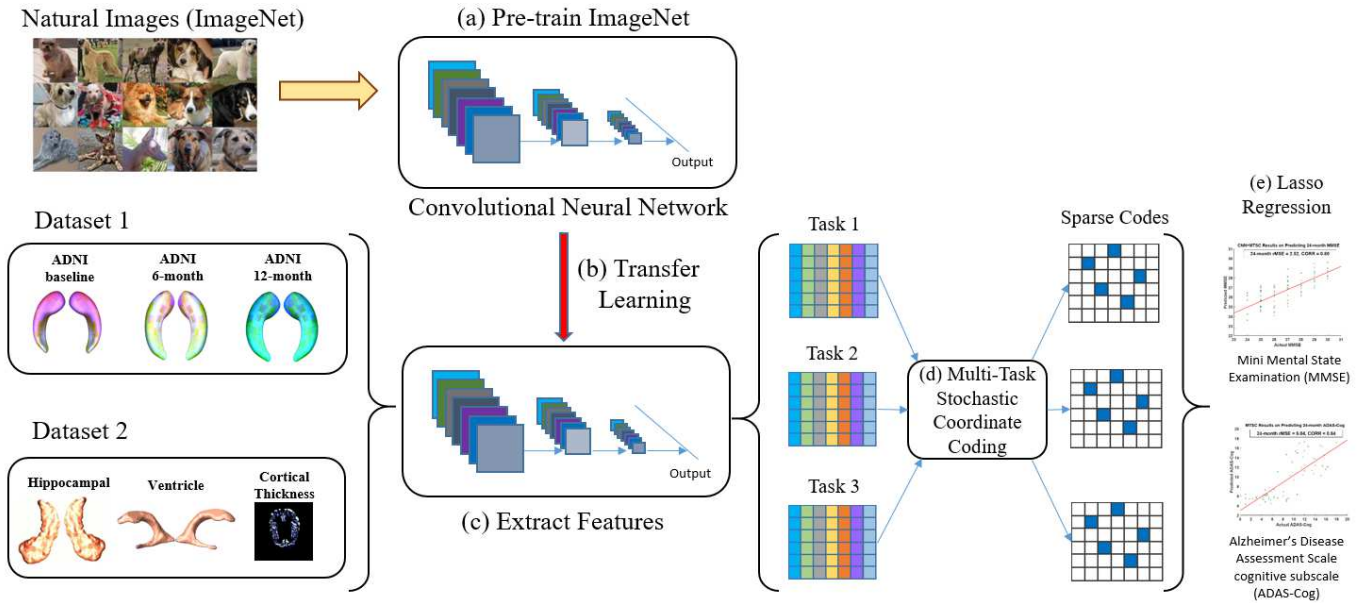}
\caption{The streamline of our proposed framework. We pre-train the deep CNN model on the Imagenet dataset and use the pre-trained model as a feature extractor for the ADNI dataset. We employ the extracted features from three time points or ROIs to conduct the multi-task dictionary learning for AD progression prediction, generating the sparse features for different time points or ROIs. Finally, we use Lasso regression on the learnt features to predict future MMSE and ADAS-Cog scores.}
\label{fig:2}
\vspace{-1em}
\end{figure*}

In fact, even when we are able to transfer knowledge from the large amounts of image data to some other domain, employing transfer learning with deep model on longitudinal or multiple brain region of interests (ROIs) data is still challenging. Fig.~\ref{fig:1} depicts three most promising imaging ROIs associated with brain image analysis~\cite{thompson:natrev10}. 
Usually the large number of longitudinal or multiple ROI features measured from limited number of subjects makes it necessary to reduce feature dimensions. Dictionary learning~\cite{mairal2009online, lin2014stochastic} has been proposed to use a small number of basis vectors termed dictionary to represent local features effectively and concisely~\cite{donoho2003optimally} and help image content analysis. However, most existing works on dictionary learning  focus on the prediction of target at a single time point~\cite{mairal2009online} or on a single ROI~\cite{zhang2016hyperbolic, zhang2016applying}.
Here, we propose a novel approach that employs dictionary learning to identify important and concise features, i.e. the knowledge learned from the large amount of natural images by developing a universal representation for medical images through a deep CNN model, which is expected to improve the performance of computer aided diagnosis and prognosis.

Recently, Multi-Task Learning (MTL)~\cite{wang2015classification,zhang2012multi,zhou2013modeling, wang2014highly} has been successfully used on regression under the different time slots. Collobert \emph{et al.}~\cite{collobert2008unified} proposed a deep neural network with MTL to solve the grammatical and semantical problems on Natural Language Processing. Zhang~\emph{et al.}~\cite{zhang2015deep} integrated transfer learning with MTL on CNN for biological image analysis. Maurer \emph{et al.}~\cite{maurer2013sparse} proposed a sparse coding model for MTL and transfer learning based on the generative methods, but it is not associated with deep learning model. The common issue for the medical imaging research is that the longitudinal features of patients among different time points or features from different ROIs will always be beneficial to study together. To further improve our dictionary learning CNN model, we propose multi-task Stochastic Coordinate Coding (MSCC) algorithm to partition the dictionaries into the common and individual parts, considering the variance of subjects from multiply time points or multiple ROIs. In this study, we focus on the longitudinal dataset of a real world application, predicting future clinical scores in Alzheimer’s disease (AD). For the same subject, it may have different representations at different time points. For the traditional CNN application, it is relatively challenging to explore that the similarity and variance of subject features among different time points. We propose MSCC to learn these different tasks simultaneously and utilize both shared and individual dictionaries to encode such consistent and changing imaging features systematically.

Our main contributions can be summarized as follows:
\begin{itemize}
\vspace{-0.5em}
  \item We employ transfer learning and CNN to explore whether the transfer learning property of CNN can be enhanced to generate features from geometry mesh of biological images since the current bottleneck for CNNs to be applied to many biological problems is the limited amount of available labeled training data. We pre-train the deep neural network on the ImageNet data and transfer the knowledge of natural images to generate the neuroimaging features for the real world application.
\vspace{-0.5em}
  \item We considered the variance of subjects from different time points or ROIs and proposed a novel unsupervised dictionary learning method, termed Multi-task Stochastic Coordinate Coding (MSCC), learning the different tasks simultaneously and utilizing shared and individual dictionary to encode both consistent and changing imaging features. To the best of our knowledge, it is the first deep model to integrate multi-task learning with dictionary learning research for brain imaging analysis.
\vspace{-0.5em}
  \item We tested our hypothesis on two different feature sets (three time points and three brain areas) to better predict the future clinical cognitive scores. Specifically, we used the multiple time points features as multiple tasks input to predict future cognitive scores. We also use multiple ROIs as multiple tasks input to predict three future time points clinical scores. Our new approach outperforms seven other state-of-the-art methods and is able to boost the performance of diagnoses ranging from cognitively unimpaired to AD.
\end{itemize}
\vspace{-0.5em}
\section{Multi-task Dictionary Learning based Convolutional Neural Network}
Our first goal here is to explore whether this transfer learning framework of CNN can be generalized to biological image studies. Specifically, we pre-train the CNN model using ImageNet~\cite{deng2009imagenet} data, containing millions of labeled natural images with thousands of categories to obtain initial parameters and subsequently generate the features on the longitudinal data for each tasks. In the experiments, we apply Alexnet \cite{krizhevsky2012imagenet}, which contains 7 layers, including convolutional layers with fixed filter sizes and different numbers of feature maps. We employ rectified non-linearity, max-pooling on each layer in our CNN model. We pretrain the CNN model on the ImageNet dataset, then remove the last fully-connected layer (this layer's outputs are the 1000 class scores for a different task like ImageNet). Finally, we treat the rest of the CNN as a fixed feature extractor for the publicly available Alzheimer's Disease Neuroimaging Initiative (ADNI) database~\cite{jack2008alzheimer}.

We further propose to use multi-task learning strategy to boost the future clinical score regression accuracy. The entire pipeline of our method is illustrated in Fig.~\ref{fig:2}. To be specific, we train the deep CNN model on the Imagenet dataset firstly. Then we employ the pretrained network as a feature extractor for the ADNI dataset from multiple time points or multiple brain ROIs. The AlexNet has a seven layer structure deep neural network. As a result, we generate seven deep output features for each time point. We further employ MSCC to conduct the multi-task learning simultaneously, generating the sparse features and dictionaries from the deep features of different time points or different brain ROIs. In MSCC, we utilize shared and individual dictionaries to encode both consistent and changing imaging features along longitudinal time points. In the end, we employ the sparse codes generated from MSCC to perform the Lasso~\cite{tibshirani1996regression} and predict the future AD progression. MSCC is one kind of online learning methods and the advantage of online learning method is to solve the cases that the size of the input data might be too large (sample size up to 2867562 in this paper) to fit into memory or the input data comes in a form of a stream.
\section{Multi-task Stochastic Coordinate Coding}
\label{chap:1}
\subsection{Dictionary Learning}
\vspace{-0.7mm}
Given a finite training set of signals $X=(x_1,..., x_n)$ where $X \in \mathbb{R}^{p\times n} $. Each $x_i$ is an image patch and $x_i \in \mathbb{R}^p$. Dictionary learning aims to learn a dictionary $D$ where $D \in \mathbb{R}^{p\times l}$ and a sparse code matrix $Z$, $Z \in \mathbb{R}^{l\times n} $. The original signals $X$ is modeled by a sparse linear combination of $D$ and $Z$ as $X\approx DZ$. Given one image patch $x_i$, we can formulate the following optimization problem:
\vspace{-1.5mm}
\begin{equation}
\min_{D\in \Psi, z_i \in \mathbb{R}^{l}} f(D, z_i)  = \frac{1}{2}||x_i-Dz_i||^2_2+\lambda||z_i||_1,
\label{eq:1}
\vspace{-1.5mm}
\end{equation}
where $\Psi = \{ D \in \mathbb{R}^{p\times l}: \forall j\in 1,...,l, ||D_j||_2 \leq 1 \}$, $D_j$ denotes the $j$th column of $D$. $\lambda$ is the positive regularization parameter. $z_i$ is the learnt sparse codes for $x_i$ and $Z=(z_1,..., z_n)$.

The optimization of Eq.\ref{eq:1} can be decomposed into an alternative learning process in the Online Dictionary Learning methods (ODL)~\cite{mairal2009online}. Given each image patch $x_i$, ODL keeps the $D$ fixed and learn $z_i$, then keep $z_i$ fixed and learn $D$. The learning process runs $\kappa$ (a fixed constant) iterations until there are no more changes on $D$ and $Z$.

\begin{figure}
\centering
\includegraphics[height=4.5cm]{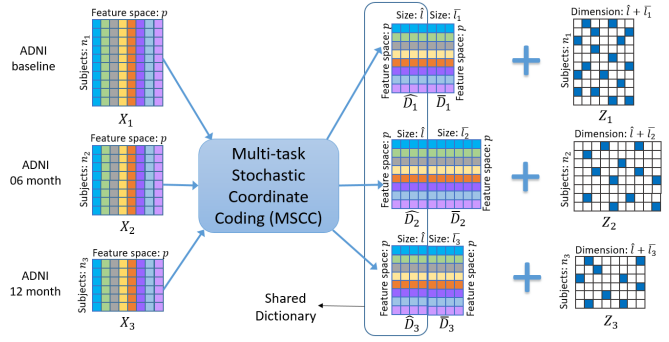}
\vspace{-0.5em}
\caption{Illustration of the learning process of MSCC.}
\vspace{-1em}
\label{fig:3}
\end{figure}

\subsection{The Proposed Algorithm}
\label{chap:2}
Given features from $T$ different tasks: $\{X_1,$$X_2$$,...,X_T\}$, our objective is to learn a set of sparse codes $\{Z_1,Z_2,...,Z_T\}$ for  each task where $X_t \in  \mathbb{R}^{p\times n_t }$, $Z_t \in  \mathbb{R}^{l_t\times n_t}$ and $t\in \{1,...,T\}$. $n_t$ is the number of subjects for $X_t$ and $l_t$ is the dimension of each sparse code in $Z_t$. When employing the ODL to learn the sparse codes $Z_t$ by $X_t$ individually, we obtain a set of dictionary $\{D_1,...,D_T\}$ but there is no correlationship between learnt dictionaries. Another solution is to construct the features $\{X_1,...,X_T\}$ into one matrix $X$ to obtain the dictionary $D$.
However, if there is no latent common information shared by the same subject during different time points, only one dictionary $D$ is not enough to show the variation among features from different time points. Such fact is supposed to be easily revealed in the variance of dictionary atoms and the sparsity of their corresponding sparse code matrices. To address this challenge, we integrate the idea of multi-task learning into the online dictionary learning method. We propose a novel dictionary learning algorithm, termed as \emph{Multi-task Stochastic Coordinate Coding} (MSCC), to learn the sparse codes of subjects from different time points.\par

For the subjects' feature matrix $X_t$ of a particular task, MSCC learns a dictionary $D_t$ and sparse codes $Z_t$. $D_t$ is composed of two parts:
$D_t = [\hat{D_t}, \bar{D_t}]$ where $\hat{D_t}\in \mathbb{R}^{p\times \hat{l}} $, $\bar{D_t}\in\mathbb{R}^ {p\times\bar{l}_t}$ and
$\hat{l}+\bar{l}_t=l_t$. $\hat{D_t}$ is the same among all the learnt dictionaries $\{D_1,...,D_T\}$ while $\bar{D_t}$ is different from each
other and only learnt from the corresponding subjects' feature matrix $X_t$. Therefore, objective function of MSCC can be reformulated as follows:
\vspace{-0.5em}
\begin{multline}
\min_{D_1,\cdots,D_T,\atop Z_1,\cdots,Z_T } \sum_{t=1}^{T}  \frac{1}{2}||X_t-[\hat{D_t},\bar{D_t}]Z_t||^2_F+\lambda  \sum_{t=1}^{T}  ||Z_t||_1 : \\ \text{ subject to } \hat{D_1} =\cdots=\hat{D_T} \text{ and } D_t \in \Psi_t
\label{eq:3}
\end{multline}
where $\Psi_t = \{ D_t \in \mathbb{R}^{p\times l_t}: \forall j\in 1,...,l_t, ||[D_t]_j||_2 \leq 1 \}$ and $[D_t]_j$ is the $j$th column of $D_t$.

\setlength{\textfloatsep}{10pt}
\begin{algorithm}[t]
\caption{Multi-task Sparse Coordinate Coding }
\begin{algorithmic}[1]
   \Require Samples from different time points: $\{X_1, X_2,..... X_T\}$ and for each $X_t$, $X_t\in  \mathbb{R}^{p\times n_t} $
   \Ensure Dictionaries and sparse codes for each time points: $\{D_1,..., D_T\}$ and $\{Z_1,..., Z_T\}$
   \For{ $k=1$ to $\kappa$ }
   \For{ $t=1$ to $T$ }
   \For{ $i=1$ to $n_t$ }
   \State Get an image patch $x_t(i)$ from sample $X_t$.
   \State  Update $\hat{D}^k_t$: $\hat{D}^k_t=\Phi$.
   \State Update $z_t^{k+1}(i)$ and index set $I_t^{k+1}(i)$ by a few steps of CCD:
   \State \quad \quad $[ z_t^{k+1}(i), I_t^{k+1}(i) ]= CCD(\hat{D}^k_t, \bar{D}^k_t, x_t(i), I_t^k(i), z_t^k(i))$.
   \State Update the $\hat{D_t}$ and $\bar{D_t}$ by one step SGD:
   \State \quad \quad $[\hat{D}^{k+1}_t,\bar{D}^{k+1}_t]= SGD(\hat{D}^k_t, \bar{D}^k_t, x_t(i), I_t^{k+1}(i), z_t^{k+1}(i) ) $.
   \State Normalize $\hat{D}^{k+1}_t$ and $\bar{D}^{k+1}_t$ based on the index set $I_t^{k+1}(i)$.
   \State Update the shared dictionary $\Phi$: $\Phi=\hat{D}^{k+1}_t$.
   \EndFor
   \EndFor
   \EndFor
\end{algorithmic}
\label{alg:2}
\end{algorithm}

Fig.~\ref{fig:3} illustrates the framework of MSCC with features of ADNI from three different time points, which represents as $X_1$, $X_2$ and $X_3$, respectively. Through the multi-task learning process of MSCC, we obtain the dictionary and sparse codes for features from each time point $t$: $D_t$ and $Z_t$. In MSCC, a dictionary $D_t$ is composed by a shared part $\hat{D}_t$ and an individual part $\bar{D}_t$, $\hat{D}_1 =$ $\hat{D}_2 =$ $\hat{D}_3$. For the individual part of dictionaries, MSCC learns a different $\bar{D}_t$ only from the corresponding feature matrix $X_t$. We vary the number of columns  $\bar{l}_t$ in $\bar{D}_t$ to introduce the variant in the learnt sparse codes $Z_t$. As a result, the dimensions of learnt sparse codes matrix $Z_t$ are different from each other.

\begin{algorithm}[t]
\caption{Updating sparse codes $z_t^{k+1}(i)$}
\begin{algorithmic}[1]
   \Require The image patch $x_t(i)$, dictionaries $\hat{D}^k_t$ and $\bar{D}^k_t$, sparse codes $z_t^k(i)$ and index set $I_t^k(i)$
   \Ensure The updated sparse code $z_t^{k+1}(i)$ and the index set $I_t^{k+1}(i)$.
   \For{ $j=1$ to $l_t$ }
   \State\quad $g = [\hat{D}^k_t,\bar{D}^k_t]_j^T (  \Omega( [ \hat{D}^k_t, \bar{D}^k_t],  z_t^k(i), I_t^k(i) ) -x_t(i) ) $
   \State\quad $ z_t^{k+1}(i)_j=\Gamma_{\lambda }( z_t^k(i)_j-g) $
   \If {$z_t^{k+1}(i)_j \neq 0$}
   \State\quad  Put $j$ into the index set $I_t^{k+1}(i)$.
   \EndIf
   \EndFor
   \For{ $s=1$ to $S$ }
   \For{ every element $\mu$ in the index set $I_t^{k+1}(i)$ }
   \State\quad $g = [\hat{D}^k_t,\bar{D}^k_t]_{\mu}^T( \Omega( [ \hat{D}^k_t, \bar{D}^k_t], z_t^{k+1}(i), I_t^{k+1}(i) )-x_t(i))$
   \State\quad $ z_t^{k+1}(i)_{\mu}=\Gamma_{\lambda }(( z_t^{k+1}(i)_{\mu}-g) $
   \EndFor
   \EndFor
\end{algorithmic}
\label{alg:3}
\end{algorithm}

The initialization of dictionaries in MSCC is critical to the entire learning process. We propose a random patch method to initialize the dictionaries from different time points. The main idea of the random patch method is to randomly select $l$ image patches from $n$ subjects $\{x_1, x_2, ..., x_n\}$ to construct $D$ where $D\in \mathbb{R}^{p\times l}$. It is a similar way to perform the random patch approach in MSCC. In MSCC, the way we initialize $\hat{D}_t$ is to randomly select $\hat{l}$ subjects' feature from features' matrices across different time points $\{X_1,\cdots,X_T\}$ to construct it. For the individual part of each dictionary, we randomly select $\bar{l}$ subjects' feature from the corresponding matrix $X_t$ to construct $\bar{D}_t$.

After initializing dictionary $D_t$ for each time point, we set all the sparse code $Z_t$ to be zero at the beginning. The key steps of MSCC are summarized in Algorithm~\ref{alg:2}.

In algorithm~\ref{alg:2}, $k$ denotes the epoch number where $k\in[1,\kappa]$. $\Phi$ represent the shared part of each dictionary $D_t$ which is initialized by the random patch method. For each subject's feature $x_t(i)$ extracted from $X_t$, we learn the $i$th sparse code  $z_t^{k+1}(i)$ from $Z_t$ by several steps of Cyclic Coordinate Descent (CCD)~\cite{canutescu2003cyclic}. Then we use learnt sparse codes $z_t^{k+1}(i)$ to update the dictionary $\hat{D}^{k+1}_t$ and $\bar{D}^{k+1}_t$ by one step Stochastic Gradient Descent (SGD)\cite{zhang2004solving}. Since $z_t^{k+1}(i)$ is very sparse, we use the index set $I_t^{k+1}(i)$ to record the location of non-zero entries in $z_t^{k+1}(i)$ to accelerate the update of sparse codes and dictionaries. $\Phi$ is updated in the end of $k$th iteration to ensure $\hat{D}^{k+1}_t$ is the same among all the dictionaries.

\begin{algorithm}[t]
\caption{Updating dictionaries $\hat{D}^{k+1}_t$ and $\bar{D}^{k+1}_t$}
\begin{algorithmic}[1]
   \Require The image patch $x_t(i)$, dictionaries $\hat{D}^k_t$ and $\bar{D}^k_t$, sparse codes $z_t^{k+1}(i)$ and index set $I_t^{k+1}(i)$.
   \Ensure The updated dictionaries $\hat{D}^{k+1}_t$ and $\bar{D}^{k+1}_t$.
   \State  Update the Hessian matrix $H_t^{k+1}$: $H_t^{k+1} = H_t^k+ z_t^{k+1}(i) z_t^{k+1}(i)^T$.
   \State   $R =  \Omega( [ \hat{D}^k_t, \bar{D}^k_t], z_t^{k+1}(i), I_t^{k+1}(i) )-x_t(i)$.
   \For{ $j=1$ to $p$ }
   \For{ every element $\mu$ in the index set $I_t^{k+1}(i)$ }
   \State\quad   $[\hat{D}^{k+1}_t,\bar{D}^{k+1}_t]_{j, \mu} = [\hat{D}^k_t,\bar{D}^k_t]_{j, \mu} - \frac{1}{H_t^{k+1}(\mu, \mu)} z_t^{k+1}(i)_{\mu} R_j$.
   \EndFor
   \EndFor
\end{algorithmic}
\label{alg:4}
\end{algorithm}

\subsection{Updating Sparse Codes and Dictionaries}
The learning process of sparse code $z_t^{k+1}(i)$ is shown in algorithm \ref{alg:3}. At first, we generate the non-zero index set $I_t^{k+1}$ by one step of CCD to record the nonzero entry of $z_t^{k+1}(i)$. Then we perform $S$ steps CCD to update the sparse codes only on the non-zero entries of $z_t^{k+1}(i)$, accelerating the learning process significantly. $\Omega$ is a sparse matrix multiplication function that has three input parameters. Take $\Omega (A, b, I)$ as an example, $A$ denotes a matrix, $b$ is a vector and $I$ is an index set that records the locations of non-zero entries in $b$. The return value of function $\Omega$ is defined as: $\Omega (A, b, I)=Ab$. When multiplying $A$ and $b$, we only manipulate the non-zero entries of $b$ and corresponding columns of $A$ based on the index set $I$, speeding up the calculation by utilizing the sparsity of $b$. $\Gamma$ is the soft thresholding shrinkage function \cite{combettes2005signal} and the definition of $\Gamma$ is  given by: $\Gamma_{\varphi}(x)=sign(x)(|x|-\varphi)$.

The procedure of updating dictionaries is shown in Algorithm \ref{alg:4}. We perform one step SGD to update the dictionaries: $\hat{D}^{k+1}_t$ and $\bar{D}^{k+1}_t$. The learning rate is set to be an approximation of the inverse of the Hessian matrix $H_t^{k+1}$, which is updated by the sparse codes $z_t^{k+1}(i)$ in $k$th iteration. For the $\mu$th column of dictionary, we set the learning rate as the inverse of the diagonal element of the Hessian matrix, which is $1/H_t^{k+1}(\mu, \mu)$. Since $D_t \in \Psi_t$ in equation (\ref{eq:3}), it is necessary to normalize the dictionaries $\hat{D}^{k+1}_t$ and $\bar{D}^{k+1}_t$ after updating them. We can perform the normalization on the corresponding columns of non-zero entries from $z_t^{k+1}(i)$ because the dictionaries updating only occurs on these columns. Utilizing the non-zero information from $I_t^{k+1}(i)$ can accelerate the whole learning process.

\vspace{-0.5em}
\section{Experiments}
AD and its early stage, Mild Cognitive Impairment (MCI), are becoming the most prevalent neurodegenerative brain diseases in elderly people worldwide~\cite{brookmeyer2007forecasting}. To this end, there have been a lot of efforts on investigating the underlying biological or neurological mechanisms and also discovering biomarkers for early diagnosis of AD and MCI. We conducted experiments from ADNI dataset~\cite{jack2008alzheimer}, which has been considered as the benchmark database for performance evaluation of various methods for AD diagnosis. We evaluated our method on two different sets of structural magnetic resonance imaging (MRI) data from the ADNI dataset: multiple time point hippocampal surface feature dataset (HP)~\cite{zhang2017multi} and multiple baseline brain ROI surface feature dataset (ROI). Specifically, for the HP dataset, we predicted clinical scores of patients at 24-month using their surface features at baseline, 6-month and 12-month. For the ROI dataset, we predicted clinical scores of patients at 6-month, 12-month and 24-month using their baseline hippocampal, ventricular and cortical thickness surface features.

\begin{table}[t]
\centering
\caption{The architecture of our CNN used in HP and ROI. }
\vspace{-0.5em}
\label{CNNarch}
 \begin{tabular}{|c|c|c|}
\hline
\captionsetup[table]{skip=10pt}
Deep Layer & Function &  $\#$ of neurons \\ \hline
1 & Convolutional Layer & 253440 \\ \hline
2 & Pooling Layer & 186624 \\ \hline
3 & Convolutional Layer & 64896 \\ \hline
4 & Convolutional Layer & 64896 \\ \hline
5 & Convolutional layer & 43264 \\
   & Pooling layer & 9216 \\ \hline
6 & Fully connected layer & 4096 \\ \hline
7 & Fully connected layer & 4096 \\ \hline
\end{tabular}
\end{table}

\subsection{Experimental Setup}
We built a prediction model for each of the above datasets using multiple task geometry surface features. To train the CNN model, patches of size 50 $\times$ 50 are extracted from surface mesh structures. We implemented our CNN model using the Caffe toolbox~\cite{jia2014caffe} and the architecture of our CNN is shown in Tab.~\ref{CNNarch}. The network was trained on a Intel (R) Xeon (R) 48-core machine, with 2.50 GHZ processors, 256 GB of globally addressable memory and a single Nvidia GeForce GTX TITAN black GPU. In the experimental setting of MSCC, the sparsity $\lambda = 0.1$. Also, we selected 10 epochs with a batch size of 1 in Algorithm~\ref{alg:2} and 3 iterations of CCD in Algorithm~\ref{alg:3} ($P$ is set to be 1 and $S$ is set to be 3) in all the experiments.   After we get the MSCC features, we used Max-Pooling~\cite{boureau2010theoretical} for further dimension reduction. Therefore, the feature dimentsion of each subject is a 1 $\times$ 2000 vector. To predict future clinical scores, we used Lasso regression. For the parameter selection, 5-fold cross validation is used to select model parameters in the training data (between $10^{-3}$ and $10^3$). We used the same method for all seven other comparison methods.

In order to evaluate the model, we randomly split the data into training and testing sets using an 8:2 ratio and used 10-fold cross validation to avoid data bias. Lastly, we evaluated the overall regression performance using normalized mean square error (nMSE), weighted correlation coefficient (wR) and root mean square error (rMSE) for task-specific regression performance measures. The three measures are defined as follows:
\begin{equation*}
\begin{split}
\vspace{-0.5em}
 nMSE(Y, \hat{Y})&=\frac{\sum_{i=1}^t||Y_i-\hat{Y_i}||_2^2/\sigma(Y_i)}{\sum_{i=1}^t n_i},\\
wR(Y, \hat{Y})&=\frac{\sum_{i=1}^tCorr(Y_i, \hat{Y_i})n_i}{\sum_{i=1}^t n_i},\\
rMSE(y, \hat{y})&=\sqrt{\frac{||y-\hat{y}||_2^2}{n}}.
\end{split}
\end{equation*}
For nMSE and wR, $Y_i$ is the ground truth of target of task $i$ and $\hat{Y_i}$ is the corresponding predicted value, $\sigma(Y_i)$ is the Standard deviation of $Y_i$, $Corr$ is the correlation coefficient between two vectors and $n_i$ is the number of subjects of task $i$. For rMSE, $y$ is the ground truth of target at a single task and $\hat{y}$ is the corresponding prediction by a prediction model. The smaller nMSE and rMSE, as well as the bigger wR mean the better results. We reported the mean and standard deviation based on 40 iterations of experiments on different splits of data.

\begin{figure}
\vspace{-1.3em}
\centering
\includegraphics[height=3cm]{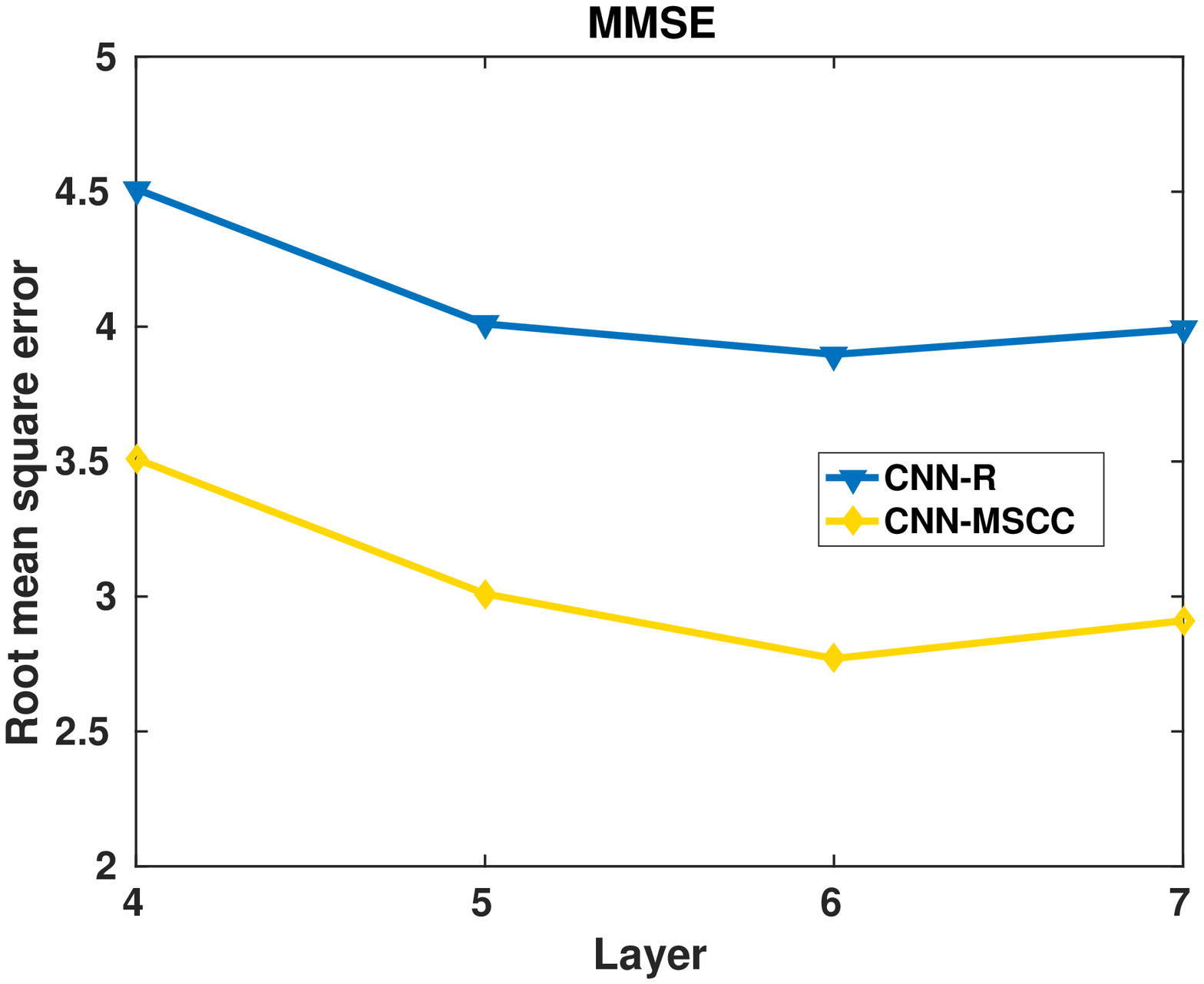}
\includegraphics[height=3cm]{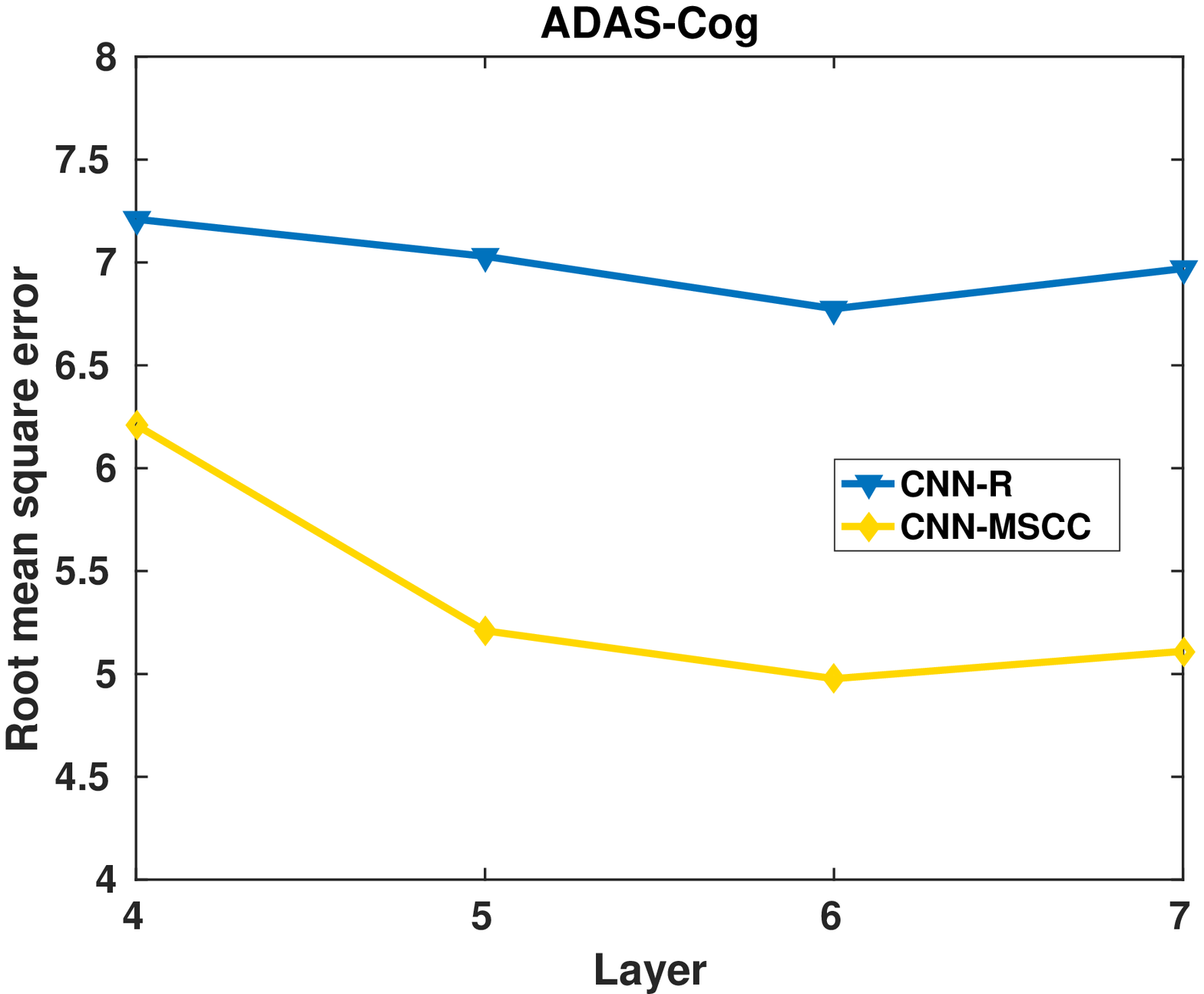}
\vspace{-0.5em}
\caption{Comparison of rMSE performance achieved by features extracted from different layers of the deep models.}
\label{fig:7}

\end{figure}

We compared the proposed model with some state-of-the-art methods, which are as follows:
\begin{itemize}
  \item CNN-MSCC: Our proposed model.
  \item CNN-R: CNN learned surface feature without transfer learning, followed by Lasso regression.
  \item MSCC-R: The proposed multi-task dictionary learning algorithm followed by Lasso regression.
  \item OLSC-R: The single-task dictionary learning~\cite{mairal2009online} followed by Lasso regression.
  \item cFSGL: A state-of-the-art multi-task algorithm called convex fused sparse group Lasso~\cite{zhou2013modeling}.
  \item L21: A state-of-the-art multi-task algorithm called $L_{2, 1}$ norm regularization with least square loss~\cite{argyriou2008convex}.
  \item Lasso: A state-of-the-art single task method called Lasso regression~\cite{tibshirani1996regression}.
  \item Ridge: A state-of-the-art single task method called Ridge regression~\cite{hoerl1970ridge}.
\end{itemize}

\subsection{Multiply Time-slots Hippocampal Surface Feature Dataset (HP)}
\label{chap:HP}
Hippocampus is a subcortical structure in the medial temporal lobe of the brain~\cite{thompson2004mapping}. Parametric shape models of the hippocampi are commonly developed for tracking shape differences or longitudinal atrophy in brain diseases. HP dataset consists of a total of 2246 subjects, consisting of 837 baseline, 733 6-month and 676 12-month imaging data. First, we used FIRST software~\cite{patenaude2011bayesian} and marching cube method~\cite{lorensen1987marching} to automatically segment and reconstruct hippocampal surfaces for each brain MR image. Then, we registered and computed surface multivariate morphometry statistics~\cite{wang2011surface}, which consist of surface multivariate tensor-based morphometry and radial distances. For each subject, we obtained a 120,000 dimensional features of the hippocampal surfaces and we use a $50\times 50$ window to obtain a collection of image patches as mentioned in Sec. 4.1. After preprocessing the data, we have 220968, 193512, 178464 image patches for different time points, respectively. Our goal is to predict Mini Mental State Examination (MMSE) and Alzheimer's Disease Assessment Scale cognitive subscale (ADAS-cog) of the 24-month patients. We used 12-month features learned by MSCC as Lasso design matrix (since it contains the baseline, 6, 12-months surface features) to train and test the 24-month clinical scores.

\begin{figure}
\vspace{-1.3em}
\centering
\includegraphics[height=3cm]{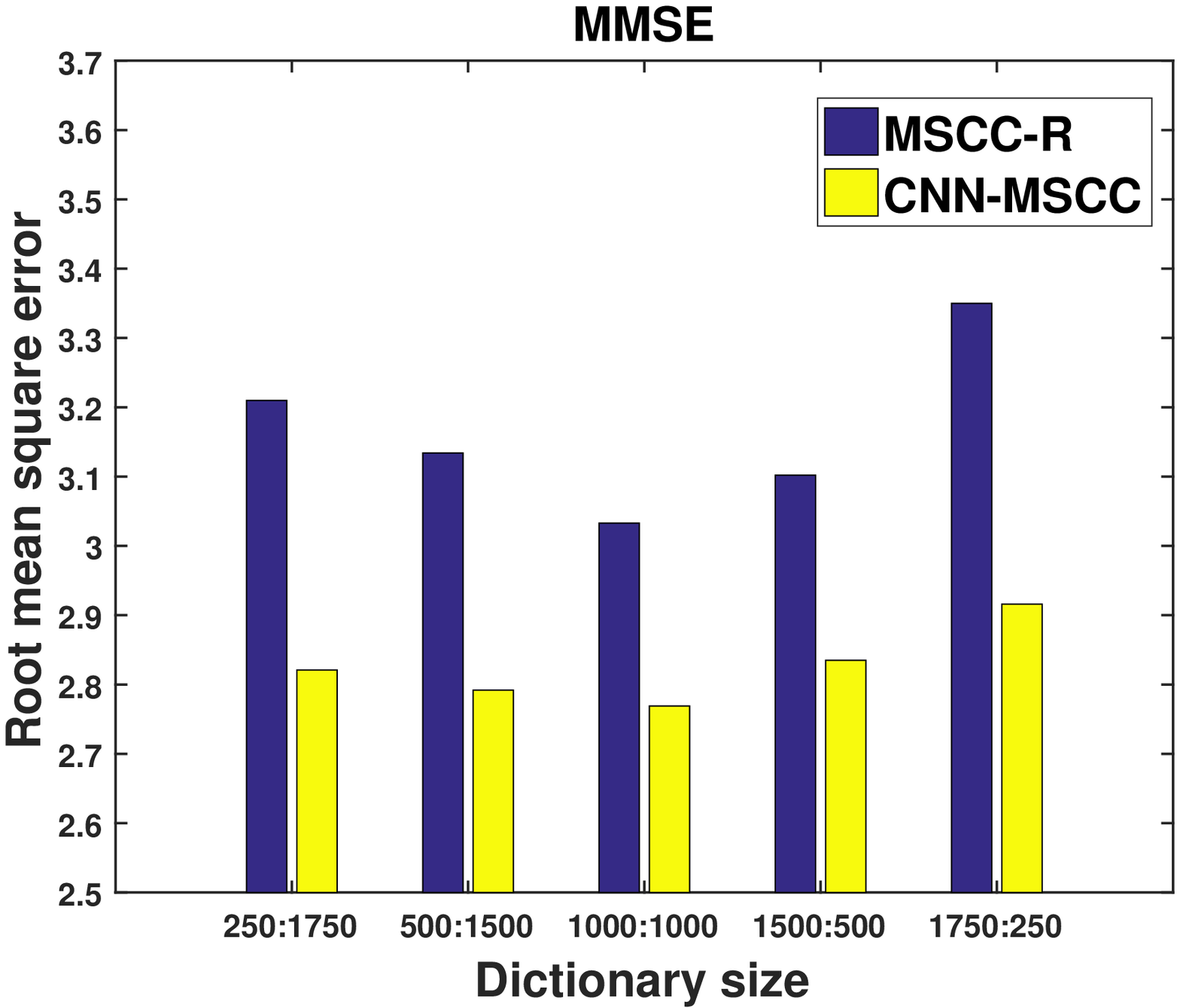}
\includegraphics[height=3cm]{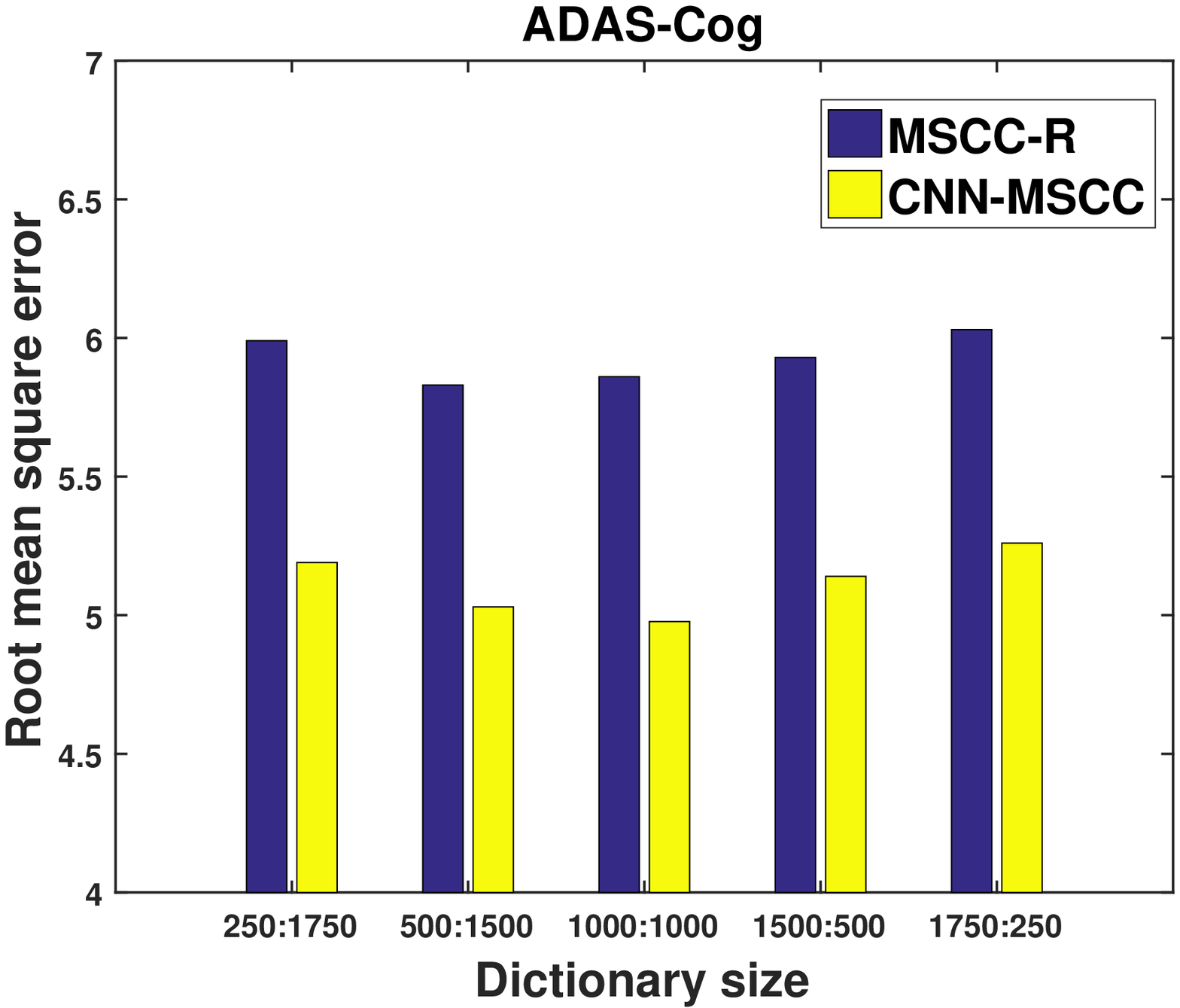}
\vspace{-0.5em}
\caption{Comparison of rMSE performance by varying the size of common dictionary on HP dataset.}
\label{fig:8}
\end{figure}

\begin{figure*}[t]
\vspace{-1.3em}
\centering
\includegraphics[height=3cm]{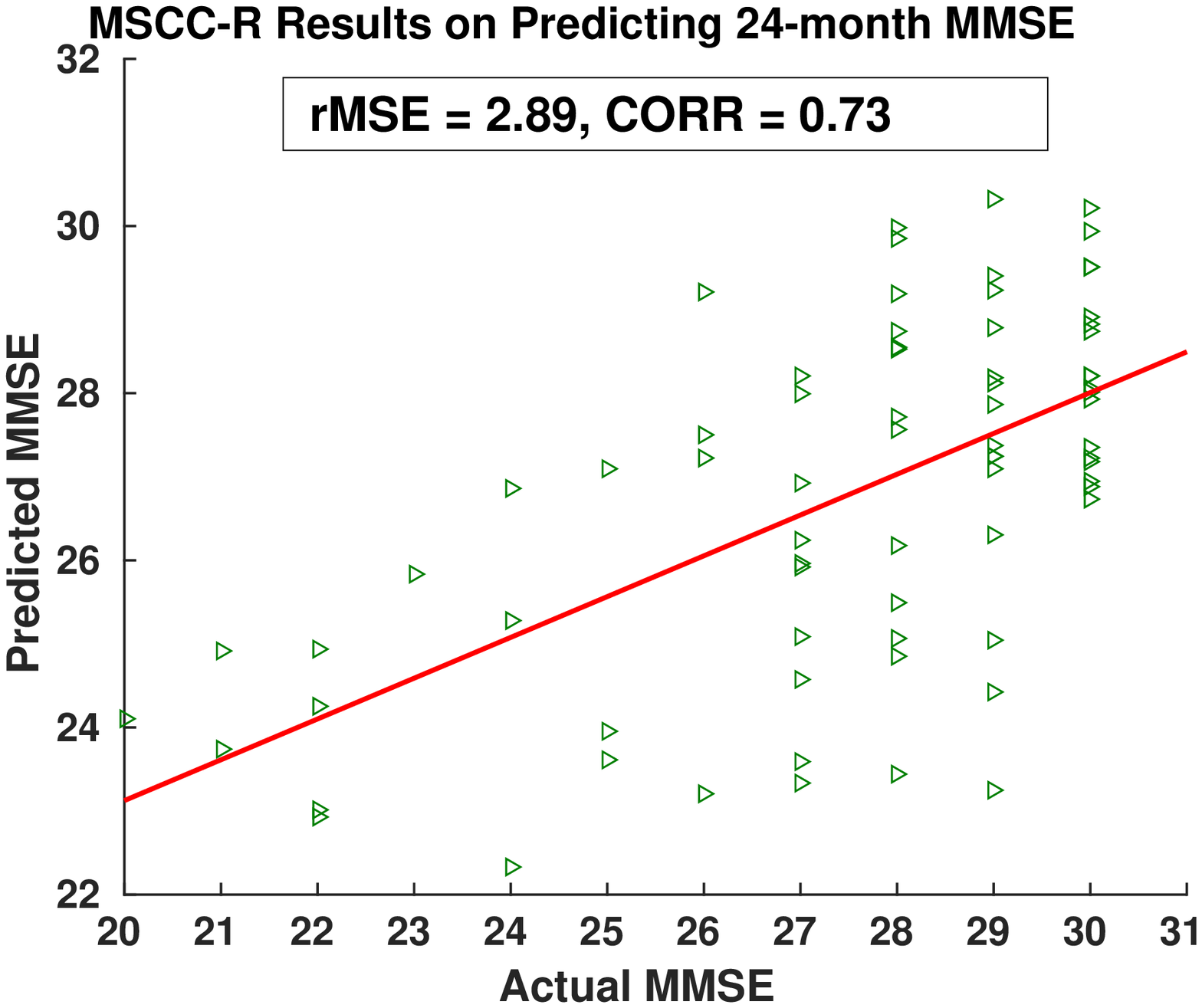}
\includegraphics[height=3cm]{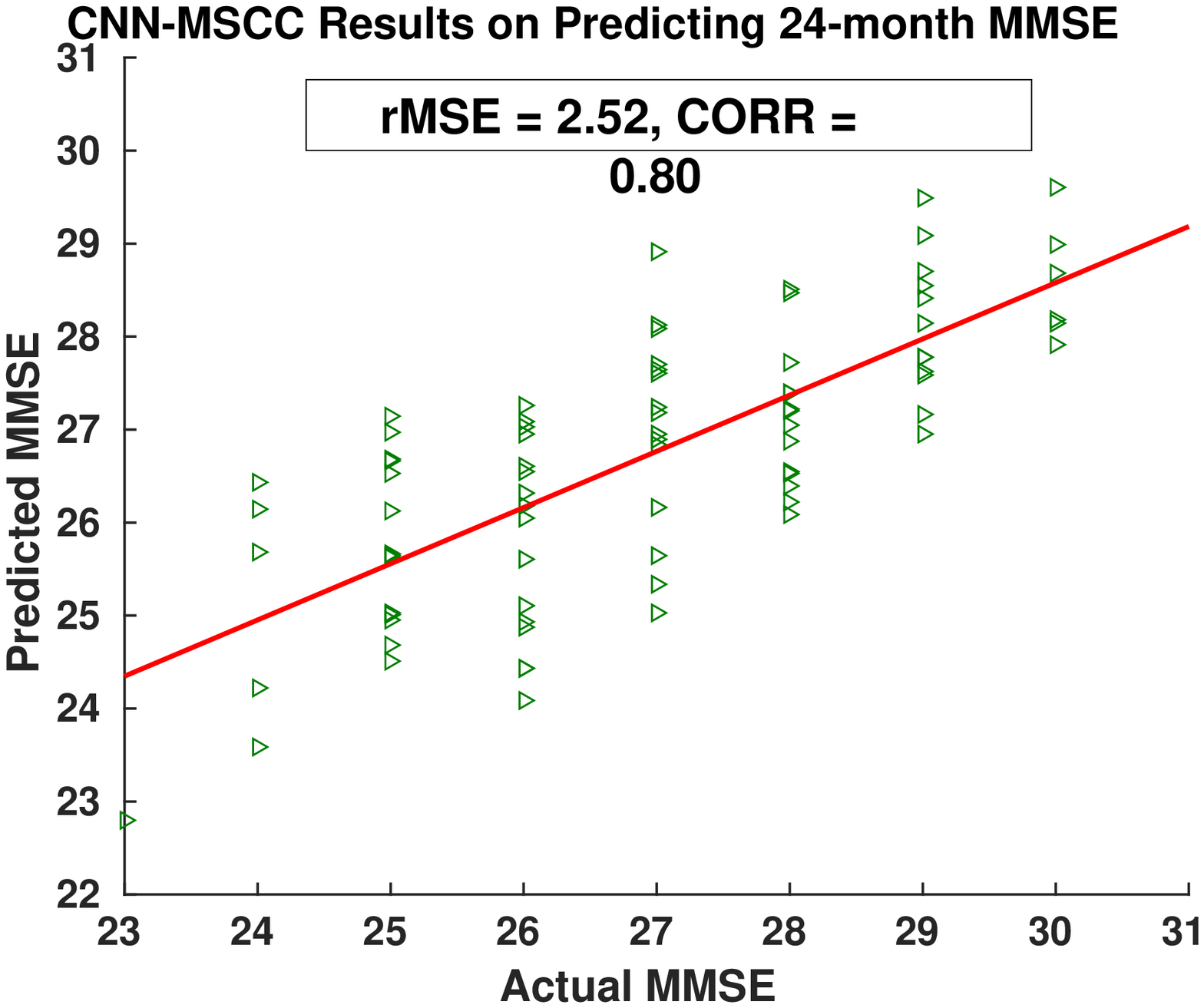}
\includegraphics[height=3cm]{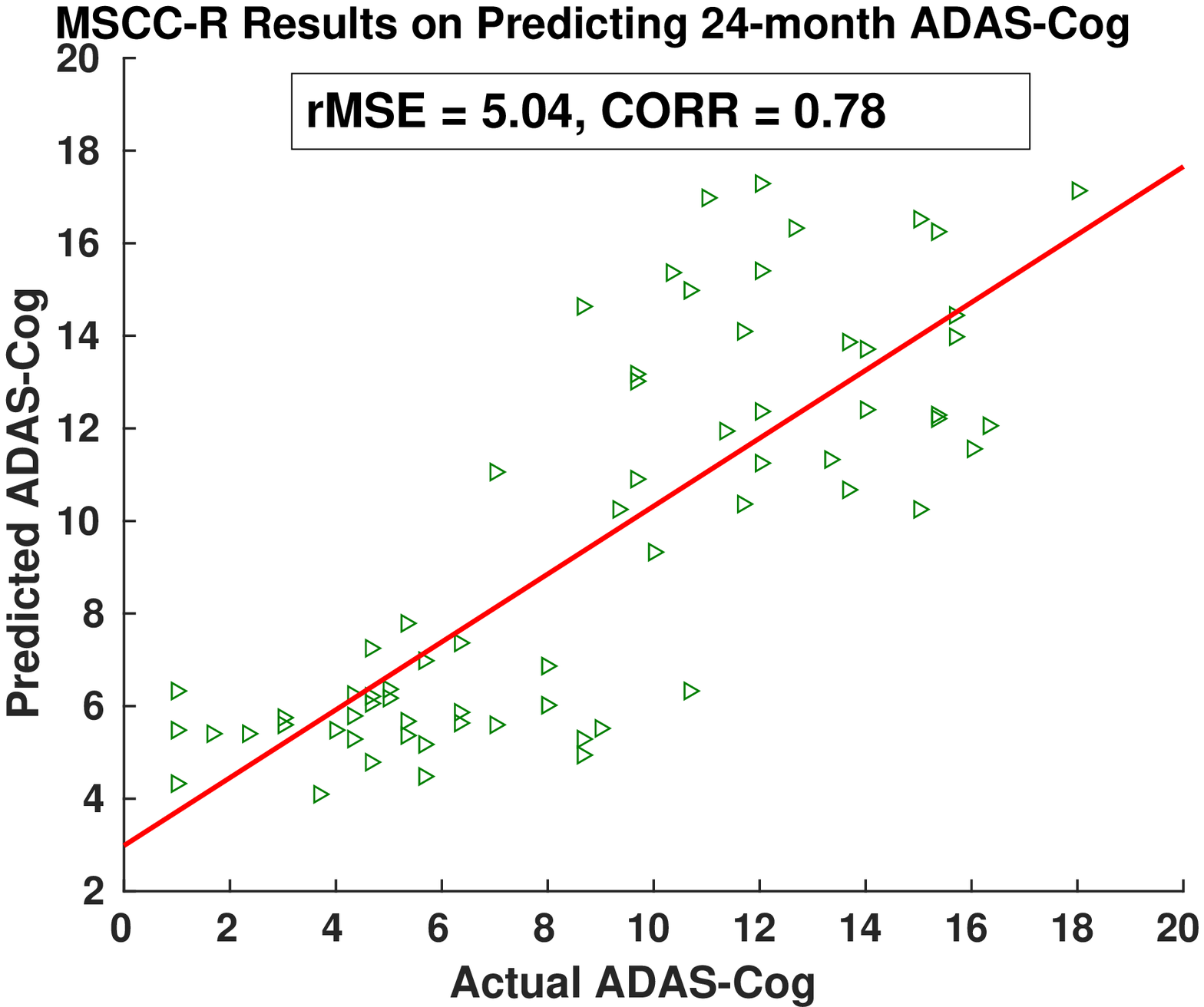}
\includegraphics[height=3cm]{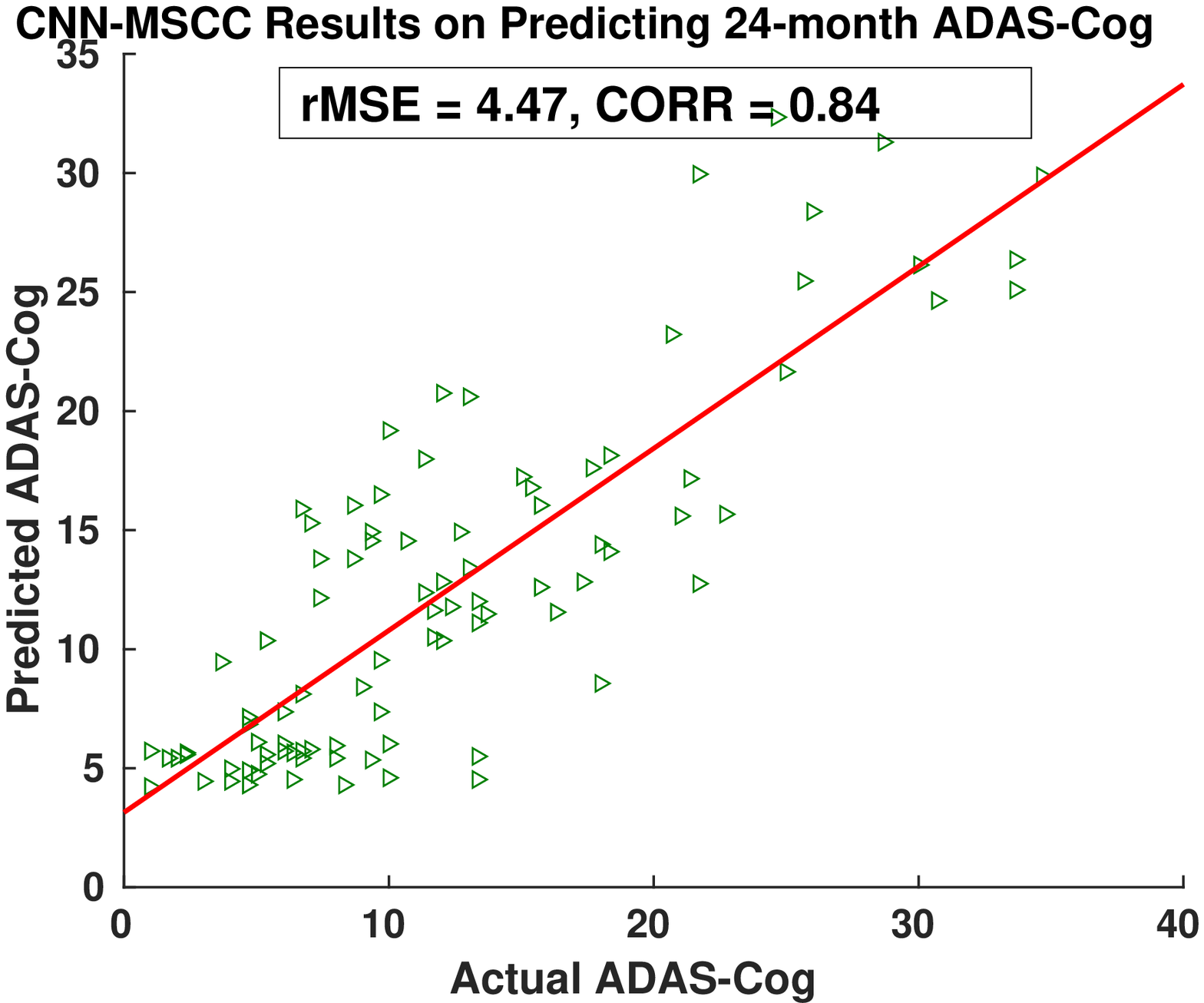}
\vspace{-0.5em}
\caption{The scatter plots for MMSE and ADAS-cog of CNN-MSCC and MSCC-R on HP dataset.}
\label{fig:4}
\vspace{-1em}
\end{figure*}

\begin{figure*}[t]
\centering
\includegraphics[height=3cm]{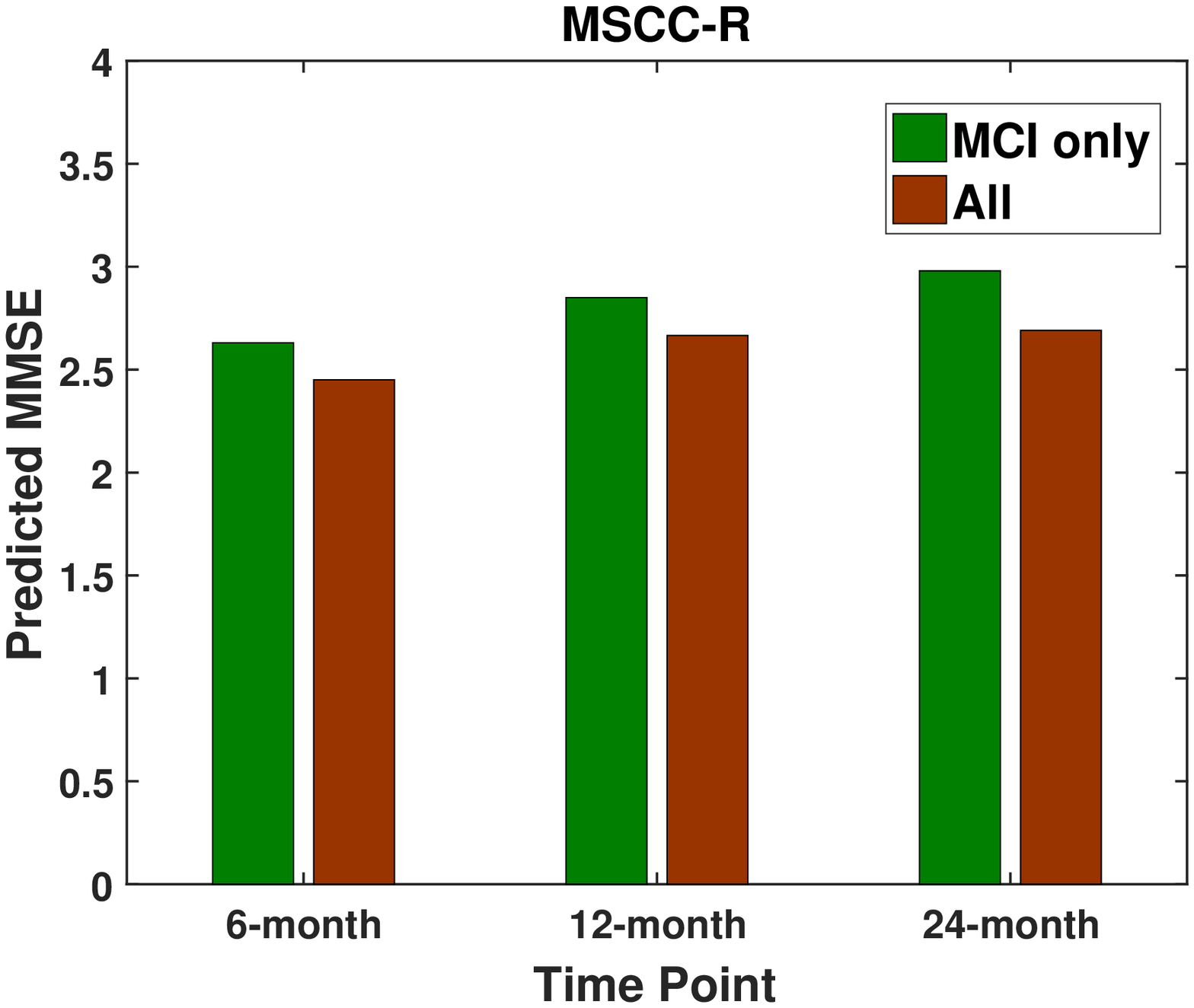}
\includegraphics[height=3cm]{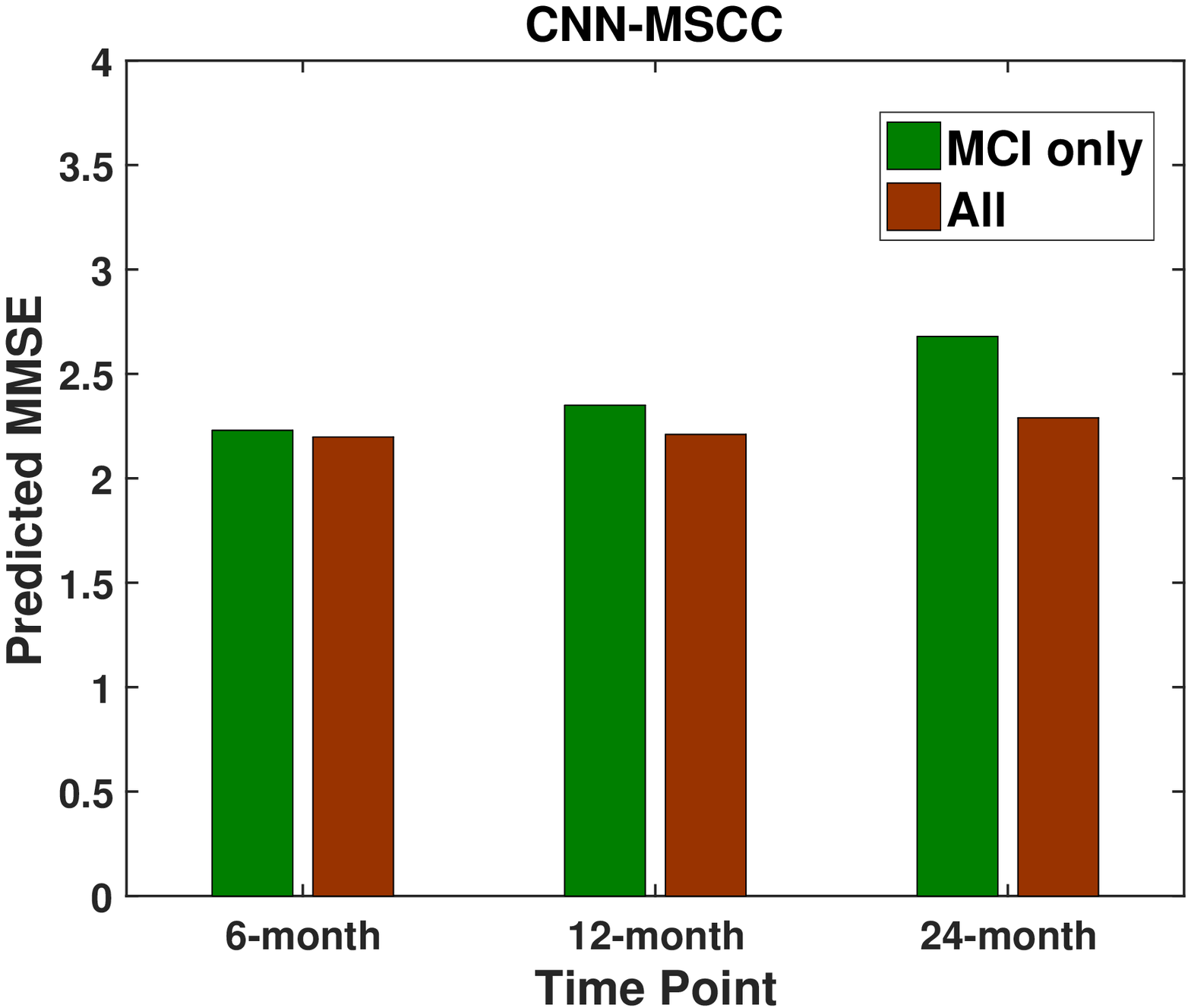}
\includegraphics[height=3cm]{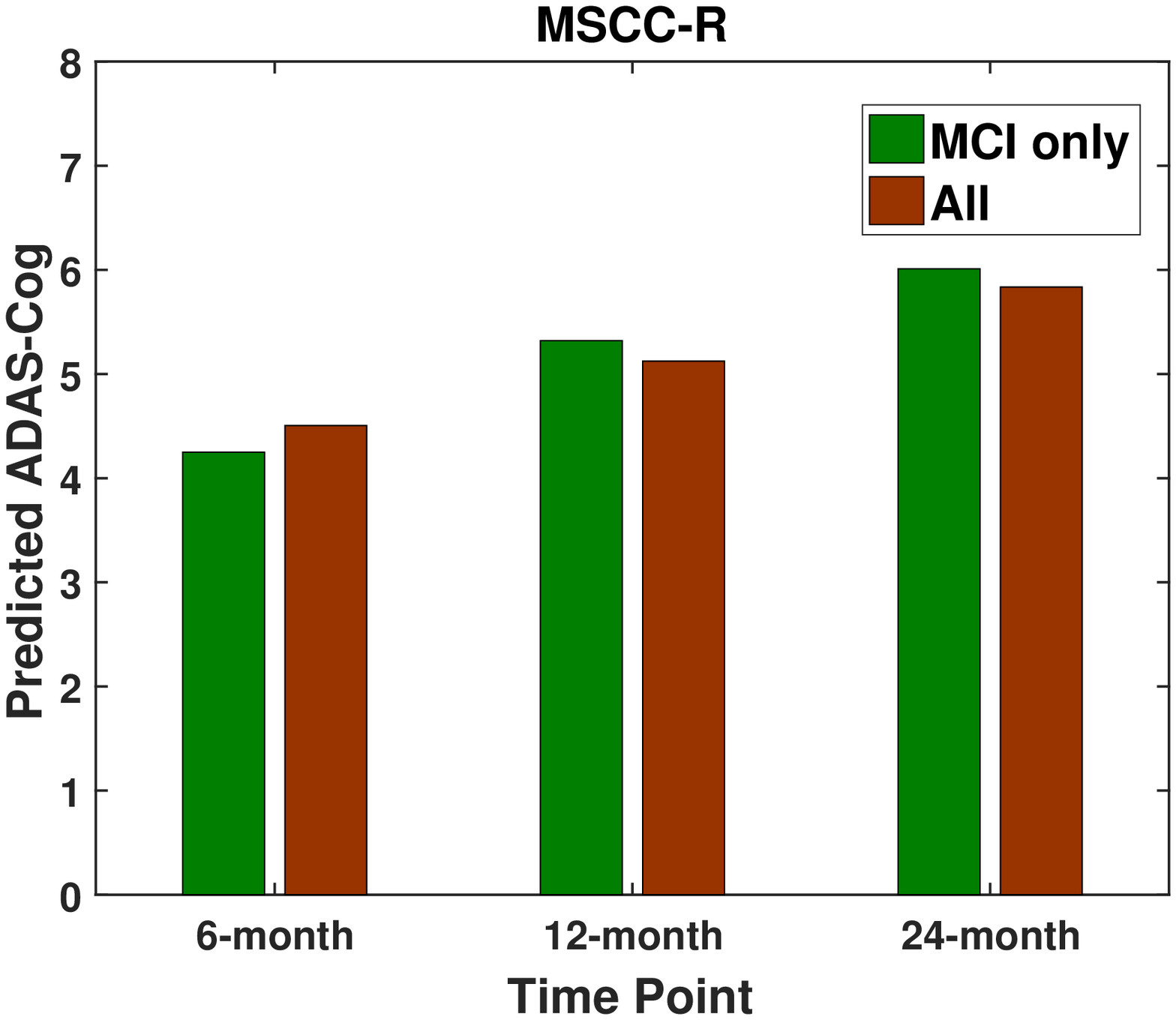}
\includegraphics[height=3cm]{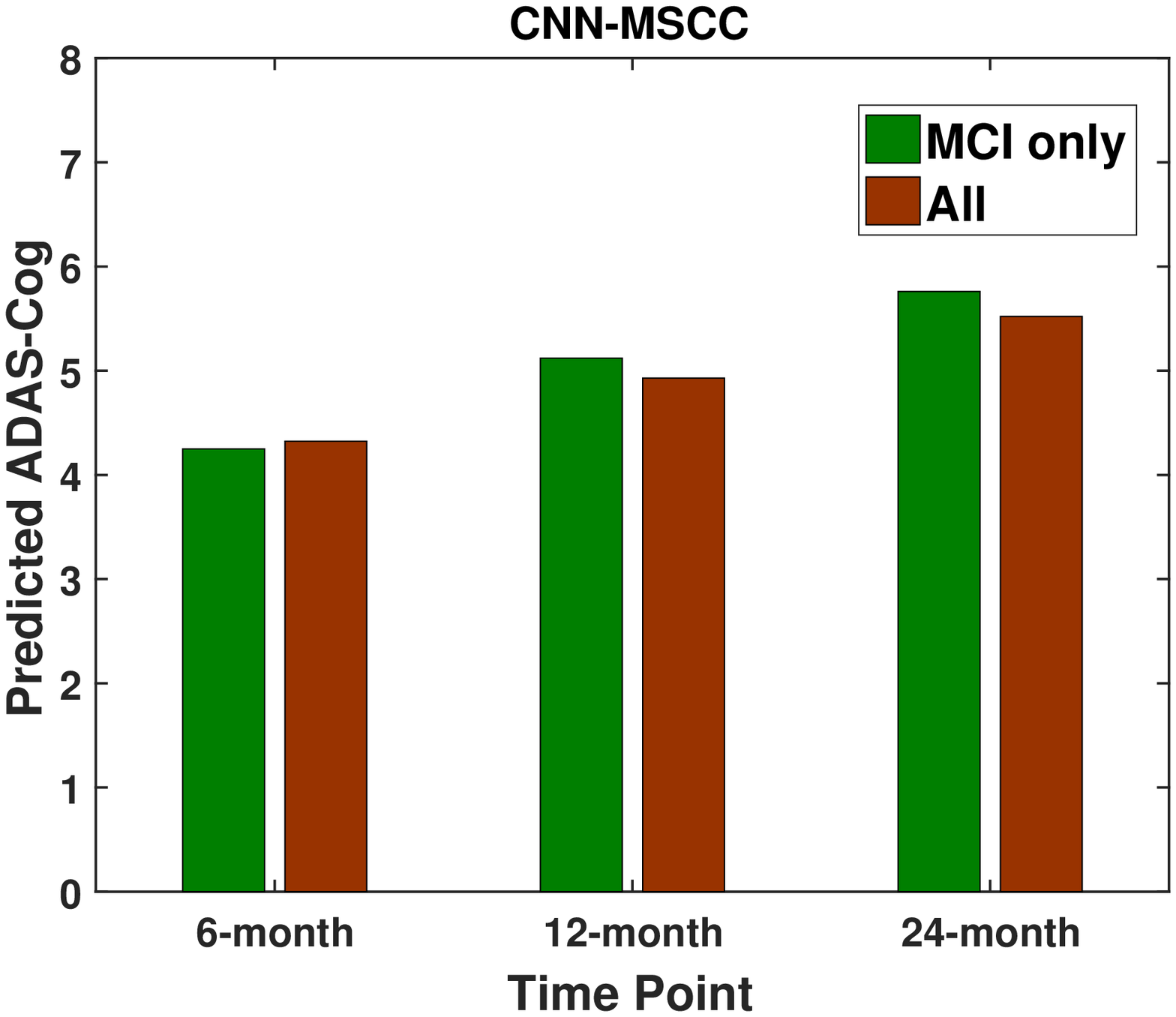}
\vspace{-0.5em}
\caption{Comparison of MMSE and ADAS-cog prediction models in terms of rMSE on patients using only MCI patients in training (MCI only), and using MCI patients together with AD patients and normal controls (All) on ROI dataset. }
\label{fig:6}
\vspace{-1em}
\end{figure*}
\vspace{0.5em}
\paragraph{Comparison of Features from Different Layers} The deep learning model consists of multiple layers of feature maps, whereby each layer is a different representation of the input data. With this hierarchical representation, we selected the layers which has the most discriminative power to capture the characteristics of the input data by comparing the features extracted from various layers of the deep model CNN-R and CNN-MSCC. Specifically, we used the HP data as inputs to train the network and extracted features from 4th, 5th, 6th and 7th layers. These features were used to predict the MMSE and ADAS-cog of 24-month patients, the results are given in Fig.~\ref{fig:7}. We observed that the 6th layer features outperformed the others in terms of overall performance in all three different deep models. The discriminative power increases from the 4th to 6th layer, and then drops afterwards as the depth of network increases. One reasonable explanation about this observation is the lower layers do not fully capture the surface features and the higher layers captured features that are specific to the training natural image set, and these features may not be relevant for surface features. In this paper, we use the 6th layer's features (4096) as the number of rows for all the dictionaries.

\paragraph{The Size of Common Dictionaries in MSCC} 
In MSCC, the common dictionary is assumed to be shared by different tasks. It is necessary to evaluate what is the appropriate size of such common dictionary. Therefore, we set the dictionary size to be 2000 and partitioned the dictionary by different proportions: 250:1750, 500:1500, 1000:1000, 1500:500 and 1750:250. The left one is the size of common dictionary while the right one is the size of individual dictionary for each task. We used two methods MSCC-R and CNN-MSCC to evaluate the regression performance. Fig.~\ref{fig:8} shows the results of rMSE of MMSE and ADAS-cog prediction on HP dataset. As it shows in Fig.~\ref{fig:8}, the rMSE of MMSE and ADAS-Cog are lowest when we split the dictionary by half and a half. It means the both of common and individual dictionaries are of equal importance during the multi-task learning. In all experiments, we use the split of 1000:1000 as the size of common and individual dictionaries and 2000 is the number of columns (dimension of each sparse code) for all the dictionaries.

\begin{table}[b]
\centering
\caption{The results of predicting 24-month MMSE and ADAS-cog on HP dataset.}
\label{tab:1}
\vspace{-0.7em}
\begin{tabular}{ccc}
\toprule
Methods& MMSE rMSE&ADAS-cog rMSE\\\hline
CNN-MSCC &\textbf{2.769$\pm$0.455}& \textbf{4.977$\pm$1.000}\\
CNN-R &3.897$\pm$0.565&6.775$\pm$1.680\\
MSCC-R&3.033$\pm$0.747&5.860$\pm$1.135\\
OLSC-R&4.656$\pm$0.575&7.553$\pm$1.022\\
cFSGL&3.635$\pm$0.707&6.132$\pm$1.889\\
L21&4.323$\pm$0.834&7.910$\pm$0.719\\
Lasso&5.113$\pm$0.723&7.024$\pm$0.880\\
Ridge&4.957$\pm$0.443&7.661$\pm$0.677\\
\bottomrule
\vspace{-2em}
\end{tabular}
\end{table}

\paragraph{Performance Comparison.}
We compared the results of CNN-MSCC with other state-of-the-art methods on predicting 24-month MMSE and ADAS-cog in Table~\ref{tab:1} and CNN-MSCC outperformed all other methods. The results of CNN-R and cFSGL are very close while MSCC-R methods are superior to them. For dictionary learning models, we observe that MSCC-R obtained a lower rMSE result than traditional dictionary learning method OLSC-R since we consider the correlation between different time slots for different tasks and the relationship with different time points on the same patient among all tasks. For the multi-task methods, we observed MSCC-R has better performance than L21 and cFSGL. Comparing with single-task methods, we noticed that the dictionary learning methods have better performance.  We also show scatter plots of CNN-MSCC and MSCC-R for the predicted values versus the actual values for MMSE and ADAS-Cog on the testing data in Fig.~\ref{fig:4}, it shows that CNN-MSCC achieved higher predictive correlation on both MMSE and ADAS-Cog.

\begin{table*}[t]
\centering
\vspace{-1.3em}
\caption{The MMSE results of 6-month, 12-month and 24-month on ROI dataset.}
\vspace{-1em}
\label{tab:2}
\begin{tabular}{cccccc}
\toprule
Methods& nMSE & wR & M06 & M12 & M24\\\hline
CNN-MSCC & \textbf{0.274$\pm$0.051} & \textbf{0.751$\pm$0.083} & \textbf{2.198$\pm$0.062} & \textbf{2.211$\pm$0.459} & \textbf{2.290$\pm$0.601}\\
CNN-R & 0.311$\pm$0.051 & 0.681$\pm$0.091 & 2.218$\pm$0.062 & 2.396$\pm$0.250 & 2.591$\pm$0.420 \\
MSCC-R& 0.308$\pm$0.058 & 0.654$\pm$0.036 & 2.451$\pm$0.357 & 2.566$\pm$0.560 & 2.859$\pm$0.494\\
OLSC-R & 0.337$\pm$0.112 & 0.692$\pm$0.074 & 2.578$\pm$0.319 & 2.954$\pm$0.746 & 3.706$\pm$0.711 \\
cFSGL& 0.312$\pm$0.037 & 0.726$\pm$0.066 & 2.424$\pm$0.315 & 2.691$\pm$0.272 & 2.906$\pm$0.907\\
L21&0.281$\pm$0.032&0.572$\pm$0.082&2.535$\pm$0.473&2.897$\pm$0.990&3.107$\pm$0.501\\
Lasso&0.302$\pm$0.078&0.423$\pm$0.073&2.659$\pm$0.804&2.904$\pm$0.658&3.335$\pm$0.692\\
Ridge&0.299$\pm$0.101&0.449$\pm$0.091&2.766$\pm$0.776&3.001$\pm$0.280&3.621$\pm$0.893\\
\bottomrule
\end{tabular}
\end{table*}

\begin{table*}[t]
\centering
\caption{The ADAS-cog results of 6-month, 12-month and 24-month on ROI dataset.}
\label{tab:3}
\vspace{-1em}
\begin{tabular}{cccccc}
\toprule
Methods & nMSE & wR & M06 & M12 & M24\\\hline
CNN-MSCC & \textbf{0.762$\pm$0.012} & \textbf{0.862$\pm$0.045} & \textbf{4.322$\pm$0.269}&\textbf{4.930$\pm$0.192}&\textbf{5.521$\pm$0.816}\\
CNN-R & 0.802$\pm$0.059 & 0.712$\pm$0.058 &5.521$\pm$0.712&5.913$\pm$0.213&6.012$\pm$0.941\\
MSCC-R&0.792$\pm$0.039&0.837$\pm$0.045&4.506$\pm$0.452&5.124$\pm$0.689&5.835$\pm$1.042\\
OLSC-R&0.828$\pm$0.079&0.681$\pm$0.052&5.080$\pm$0.589&5.860$\pm$0.608&6.179$\pm$1.001\\
cFSGL&0.795$\pm$0.052&0.836$\pm$0.031&4.451$\pm$0.340&5.230$\pm$0.589&6.249$\pm$0.996\\
L21&0.811$\pm$0.080&0.554$\pm$0.062&4.476$\pm$0.931&5.453$\pm$0.392&6.279$\pm$1.232\\
Lasso&0.809$\pm$0.110&0.518$\pm$0.080&5.295$\pm$0.763&5.799$\pm$1.001&6.729$\pm$0.705\\
Ridge&0.819$\pm$0.108&0.497$\pm$0.071&5.534$\pm$0.542&5.907$\pm$0.885&6.543$\pm$0.844\\
\bottomrule
\vspace{-2em}
\end{tabular}
\end{table*}
\subsection{Multiple Baseline Brain ROIs Surface Features Dataset (ROI)}
In this experiment, we utilized three structural measures of brain, which are hippocampi (as we mentioned in Sec.~\ref{chap:HP}), lateral ventricle and cortical thickness, from the ADNI baseline dataset (N = 837). In brief, the lateral ventricles are often enlarged in disease and can provide sensitive measures of disease progression~\cite{thompson2004mapping} and the cortical thickness can be used as an anatomical index for quantifying cortical shape variations~\cite{chung2005unified}. For the hippocampal surface features, we used the same methods as HP dataset while for the ventricular surface features we did the following. First, we segmented images of the lateral ventricles to build the ventricular structure surface models using a level-set based topology preserving method~\cite{han2003topology}. Then we computed surface registrations using the canonical holomorphic one-form segmentation method~\cite{wang2009multivariate}. Finally, surface multivariate morphometry statistics~\cite{wang2011surface} were computed and obtained as a 308,247 dimensional features of the ventricular surfaces for each subject. The cortical thickness was computed by FreeSurfer~\cite{fischl2012freesurfer} which deforms the white surface to pial surface and measures deforming distance as the cortical thickness. The spherical parameter surface and weighted spherical harmonic representation~\cite{Chung:tmi08}\cite{zhang2017empowering} are used to register pial surfaces across subjects, which means each subjects have the same dimension (161,800) cortical thickness. The image patch size is $50\times 50$ as mentioned in Sec. 4.1. After preprocessing the data, we have 220968, 2867562, 1504926 image patches for multiple input tasks, respectively.

\paragraph{Performance Comparison.}
We constructed the prediction models by first forming the final baseline data from the combined three tasks features. Then, we used Lasso to individually predict 6-month, 12-month and 24-month MMSE and ADAS-cog scores with 8:2 ratio on training and testing data sets. The prediction results are reported in Table~\ref{tab:2} and Table~\ref{tab:3}. We can observe that the performance of predicting 6-month, 12-month and 24-month scores of MMSE and ADAS-Cog are improved by CNN-MSCC and MSCC-R for all three time points. We can also notice that the significant improvement of the proposed CNN-MSCC and MSCC-R for later time points (12, 24-month). This may be due to the data sparseness in later time points, as the proposed sparsity-inducing models are expected to achieve better prediction performance. Also, the improvement of ADAS-cog is more significant than MMSE.

\paragraph{Comparing MCI vs. all Baseline Dataset.}
In the study of AD, MCI patients are of particular interest because people with MCI are at high risk of progression to dementia. We studied the prediction performance on MCI patients and MCI patients together with AD patients and normal controls (CN) on ROI dataset. In the first experiment, we used only MCI patients in both training and testing data. We random split the MCI patients with 8:2 ratio for training and testing. For another experiment, we follow the same practice as in our previous experiments. The performance of predicting MMSE and ADAS-cog at all time points is given in Fig.~\ref{fig:6}. We see that in most cases the prediction performance together with AD and CN induce the performance improvement. This may be due to the small sample size at later time points, in which the information from AD and CN subjects may be useful during the learning. Our discovery may shed new light onto the clinical cognitive score prediction of AD.
\vspace{-1.5em}
\section{Conclusions and Future Work}
\vspace{-0.5em}
In this work, we proposed a deep learning model, multi-task dictionary learning based CNN to incorporate multiple time slots or multiple brain ROI imaging features, for predicting the AD clinical score. The proposed model is validated by extensive experimental studies and shown to be more efficient than seven other state-of-the-art methods. In future work, we will optimize our method and investigate its capability on brain multimodality imaging datasets.

{\small
\bibliographystyle{ieee}
\bibliography{egbib}

\begin{thebibliography}{10}\itemsep=-1pt

\bibitem{argyriou2008convex}
A.~Argyriou, T.~Evgeniou, and M.~Pontil.
\newblock Convex multi-task feature learning.
\newblock {\em Machine Learning}, 73(3):243--272, 2008.

\bibitem{blitzer2006domain}
J.~Blitzer, R.~McDonald, and F.~Pereira.
\newblock Domain adaptation with structural correspondence learning.
\newblock In {\em Proceedings of the 2006 conference on empirical methods in
  natural language processing}, pages 120--128. Association for Computational
  Linguistics, 2006.

\bibitem{boureau2010theoretical}
Y.-L. Boureau, J.~Ponce, and Y.~LeCun.
\newblock A theoretical analysis of feature pooling in visual recognition.
\newblock In {\em Proceedings of the 27th international conference on machine
  learning (ICML-10)}, pages 111--118, 2010.

\bibitem{brookmeyer2007forecasting}
R.~Brookmeyer, E.~Johnson, K.~Ziegler-Graham, and H.~M. Arrighi.
\newblock Forecasting the global burden of {A}lzheimer's disease.
\newblock {\em Alzheimer's \& dementia}, 3(3):186--191, 2007.

\bibitem{canutescu2003cyclic}
A.~A. Canutescu and R.~L. Dunbrack.
\newblock Cyclic coordinate descent: A robotics algorithm for protein loop
  closure.
\newblock {\em Protein science}, 12(5):963--972, 2003.

\bibitem{Chung:tmi08}
M.~K. Chung, K.~M. Dalton, and R.~J. Davidson.
\newblock {{T}ensor-based cortical surface morphometry via weighted spherical
  harmonic representation}.
\newblock {\em IEEE Trans Med Imaging}, 27(8):1143--1151, Aug 2008.

\bibitem{chung2005unified}
M.~K. Chung, S.~Robbins, and A.~C. Evans.
\newblock Unified statistical approach to cortical thickness analysis.
\newblock In {\em Biennial International Conference on Information Processing
  in Medical Imaging}, pages 627--638. Springer, 2005.

\bibitem{collobert2008unified}
R.~Collobert and J.~Weston.
\newblock A unified architecture for natural language processing: Deep neural
  networks with multitask learning.
\newblock In {\em Proceedings of the 25th international conference on Machine
  learning}, pages 160--167. ACM, 2008.

\bibitem{combettes2005signal}
P.~L. Combettes and V.~R. Wajs.
\newblock Signal recovery by proximal forward-backward splitting.
\newblock {\em Multiscale Modeling \& Simulation}, 4(4):1168--1200, 2005.

\bibitem{deng2009imagenet}
J.~Deng, W.~Dong, R.~Socher, L.-J. Li, K.~Li, and L.~Fei-Fei.
\newblock Imagenet: A large-scale hierarchical image database.
\newblock In {\em Computer Vision and Pattern Recognition, 2009. CVPR 2009.
  IEEE Conference on}, pages 248--255. IEEE, 2009.

\bibitem{donoho2003optimally}
D.~L. Donoho and M.~Elad.
\newblock Optimally sparse representation in general (nonorthogonal)
  dictionaries via l1 minimization.
\newblock {\em Proceedings of the National Academy of Sciences},
  100(5):2197--2202, 2003.

\bibitem{fischl2012freesurfer}
B.~Fischl.
\newblock Freesurfer.
\newblock {\em Neuroimage}, 62(2):774--781, 2012.

\bibitem{thompson:natrev10}
G.~B. Frisoni, N.~C. Fox, C.~R. Jack, P.~Scheltens, and P.~M. Thompson.
\newblock {{T}he clinical use of structural {M}{R}{I} in {A}lzheimer disease}.
\newblock {\em Nat Rev Neurol}, 6(2):67--77, Feb 2010.

\bibitem{han2003topology}
X.~Han, C.~Xu, and J.~L. Prince.
\newblock A topology preserving level set method for geometric deformable
  models.
\newblock {\em IEEE Transactions on Pattern Analysis and Machine Intelligence},
  25(6):755--768, 2003.

\bibitem{hazlett2017early}
H.~C. Hazlett, H.~Gu, B.~C. Munsell, S.~H. Kim, M.~Styner, J.~J. Wolff, J.~T.
  Elison, M.~R. Swanson, H.~Zhu, K.~N. Botteron, et~al.
\newblock Early brain development in infants at high risk for autism spectrum
  disorder.
\newblock {\em Nature}, 542(7641):348--351, 2017.

\bibitem{hoerl1970ridge}
A.~E. Hoerl and R.~W. Kennard.
\newblock Ridge regression: Biased estimation for nonorthogonal problems.
\newblock {\em Technometrics}, 12(1):55--67, 1970.

\bibitem{jack2008alzheimer}
C.~R. Jack, M.~A. Bernstein, N.~C. Fox, P.~Thompson, G.~Alexander, D.~Harvey,
  B.~Borowski, P.~J. Britson, J.~L~Whitwell, C.~Ward, et~al.
\newblock The alzheimer's disease neuroimaging initiative (adni): Mri methods.
\newblock {\em Journal of magnetic resonance imaging}, 27(4):685--691, 2008.

\bibitem{jia2014caffe}
Y.~Jia, E.~Shelhamer, J.~Donahue, S.~Karayev, J.~Long, R.~Girshick,
  S.~Guadarrama, and T.~Darrell.
\newblock Caffe: Convolutional architecture for fast feature embedding.
\newblock {\em arXiv preprint arXiv:1408.5093}, 2014.

\bibitem{krizhevsky2012imagenet}
A.~Krizhevsky, I.~Sutskever, and G.~E. Hinton.
\newblock Imagenet classification with deep convolutional neural networks.
\newblock In {\em Advances in neural information processing systems}, pages
  1097--1105, 2012.

\bibitem{lecun1998gradient}
Y.~LeCun, L.~Bottou, Y.~Bengio, and P.~Haffner.
\newblock Gradient-based learning applied to document recognition.
\newblock {\em Proceedings of the IEEE}, 86(11):2278--2324, 1998.

\bibitem{lin2014stochastic}
B.~Lin, Q.~Li, Q.~Sun, M.-J. Lai, I.~Davidson, W.~Fan, and J.~Ye.
\newblock Stochastic coordinate coding and its application for drosophila gene
  expression pattern annotation.
\newblock {\em arXiv preprint arXiv:1407.8147}, 2014.

\bibitem{lorensen1987marching}
W.~E. Lorensen and H.~E. Cline.
\newblock Marching cubes: A high resolution 3d surface construction algorithm.
\newblock In {\em ACM siggraph computer graphics}, volume~21, pages 163--169.
  ACM, 1987.

\bibitem{mairal2009online}
J.~Mairal, F.~Bach, J.~Ponce, and G.~Sapiro.
\newblock Online dictionary learning for sparse coding.
\newblock In {\em Proceedings of the 26th Annual International Conference on
  Machine Learning}, pages 689--696. ACM, 2009.

\bibitem{maurer2013sparse}
A.~Maurer, M.~Pontil, and B.~Romera{-}Paredes.
\newblock Sparse coding for multitask and transfer learning.
\newblock In {\em Proceedings of the 30th International Conference on Machine
  Learning, {ICML} 2013, Atlanta, GA, USA, 16-21 June 2013}, pages 343--351,
  2013.

\bibitem{pan2010survey}
S.~J. Pan and Q.~Yang.
\newblock A survey on transfer learning.
\newblock {\em IEEE Transactions on knowledge and data engineering},
  22(10):1345--1359, 2010.

\bibitem{patenaude2011bayesian}
B.~Patenaude, S.~M. Smith, D.~N. Kennedy, and M.~Jenkinson.
\newblock A bayesian model of shape and appearance for subcortical brain
  segmentation.
\newblock {\em Neuroimage}, 56(3):907--922, 2011.

\bibitem{sharif2014cnn}
A.~Sharif~Razavian, H.~Azizpour, J.~Sullivan, and S.~Carlsson.
\newblock Cnn features off-the-shelf: an astounding baseline for recognition.
\newblock In {\em Proceedings of the IEEE Conference on Computer Vision and
  Pattern Recognition Workshops}, pages 806--813, 2014.

\bibitem{thompson2004mapping}
P.~M. Thompson, K.~M. Hayashi, G.~I. De~Zubicaray, A.~L. Janke, S.~E. Rose,
  J.~Semple, M.~S. Hong, D.~H. Herman, D.~Gravano, D.~M. Doddrell, et~al.
\newblock Mapping hippocampal and ventricular change in alzheimer disease.
\newblock {\em Neuroimage}, 22(4):1754--1766, 2004.

\bibitem{tibshirani1996regression}
R.~Tibshirani.
\newblock Regression shrinkage and selection via the lasso.
\newblock {\em Journal of the Royal Statistical Society. Series B
  (Methodological)}, pages 267--288, 1996.

\bibitem{turaga2010convolutional}
S.~C. Turaga, J.~F. Murray, V.~Jain, F.~Roth, M.~Helmstaedter, K.~Briggman,
  W.~Denk, and H.~S. Seung.
\newblock Convolutional networks can learn to generate affinity graphs for
  image segmentation.
\newblock {\em Neural computation}, 22(2):511--538, 2010.

\bibitem{wang2014highly}
J.~Wang, Q.~Li, S.~Yang, W.~Fan, P.~Wonka, and J.~Ye.
\newblock A highly scalable parallel algorithm for isotropic total variation
  models.
\newblock In {\em Proceedings of the 31st International Conference on Machine
  Learning (ICML-14)}, pages 235--243, 2014.

\bibitem{wang2015classification}
X.~Wang, T.~Zhang, T.~M. Chaim, M.~V. Zanetti, and C.~Davatzikos.
\newblock Classification of mri under the presence of disease heterogeneity
  using multi-task learning: Application to bipolar disorder.
\newblock In {\em Medical Image Computing and Computer-Assisted
  Intervention--MICCAI 2015}, pages 125--132. Springer, 2015.

\bibitem{wang2009multivariate}
Y.~Wang, T.~F. Chan, A.~W. Toga, and P.~M. Thompson.
\newblock Multivariate tensor-based brain anatomical surface morphometry via
  holomorphic one-forms.
\newblock In {\em International Conference on Medical Image Computing and
  Computer-Assisted Intervention}, pages 337--344. Springer, 2009.

\bibitem{wang2011surface}
Y.~Wang, Y.~Song, P.~Rajagopalan, T.~An, K.~Liu, Y.~Y. Chou, B.~Gutman, A.~W.
  Toga, and P.~M. Thompson.
\newblock {{S}urface-based {T}{B}{M} boosts power to detect disease effects on
  the brain: an {N}=804 {A}{D}{N}{I} study}.
\newblock {\em Neuroimage}, 56(4):1993--2010, Jun 2011.

\bibitem{zeiler2014visualizing}
M.~D. Zeiler and R.~Fergus.
\newblock Visualizing and understanding convolutional networks.
\newblock In {\em European Conference on Computer Vision}, pages 818--833.
  Springer, 2014.

\bibitem{zhang2012multi}
D.~Zhang, D.~Shen, A.~D.~N. Initiative, et~al.
\newblock Multi-modal multi-task learning for joint prediction of multiple
  regression and classification variables in alzheimer's disease.
\newblock {\em NeuroImage}, 59(2):895--907, 2012.

\bibitem{zhang2011deep}
J.~Zhang.
\newblock Deep transfer learning via restricted boltzmann machine for document
  classification.
\newblock In {\em Machine Learning and Applications and Workshops (ICMLA), 2011
  10th International Conference on}, volume~1, pages 323--326. IEEE, 2011.

\bibitem{zhang2017empowering}
J.~Zhang, Y.~Fan, Q.~Li, P.~M. Thompson, J.~Ye, and Y.~Wang.
\newblock Empowering cortical thickness measures in clinical diagnosis of
  alzheimer's disease with spherical sparse coding.
\newblock In {\em Biomedical Imaging (ISBI 2017), 2017 IEEE 14th International
  Symposium on}, pages 446--450. IEEE, 2017.

\bibitem{zhang2017multi}
J.~Zhang, Q.~Li, R.~J. Caselli, P.~M. Thompson, J.~Ye, and Y.~Wang.
\newblock Multi-source multi-target dictionary learning for prediction of
  cognitive decline.
\newblock In {\em International Conference on Information Processing in Medical
  Imaging}, pages 184--197. Springer, 2017.

\bibitem{zhang2016hyperbolic}
J.~Zhang, J.~Shi, C.~Stonnington, Q.~Li, B.~A. Gutman, K.~Chen, E.~M. Reiman,
  R.~Caselli, P.~M. Thompson, J.~Ye, et~al.
\newblock Hyperbolic space sparse coding with its application on prediction of
  alzheimer�s disease in mild cognitive impairment.
\newblock In {\em International Conference on Medical Image Computing and
  Computer-Assisted Intervention}, pages 326--334. Springer, 2016.

\bibitem{zhang2016applying}
J.~Zhang, C.~Stonnington, Q.~Li, J.~Shi, R.~J. Bauer, B.~A. Gutman, K.~Chen,
  E.~M. Reiman, P.~M. Thompson, J.~Ye, et~al.
\newblock Applying sparse coding to surface multivariate tensor-based
  morphometry to predict future cognitive decline.
\newblock In {\em Biomedical Imaging (ISBI), 2016 IEEE 13th International
  Symposium on}, pages 646--650. IEEE, 2016.

\bibitem{zhang2004solving}
T.~Zhang.
\newblock Solving large scale linear prediction problems using stochastic
  gradient descent algorithms.
\newblock In {\em Proceedings of the twenty-first international conference on
  Machine learning}, page 116. ACM, 2004.

\bibitem{zhang2015deep}
W.~Zhang, R.~Li, T.~Zeng, Q.~Sun, S.~Kumar, J.~Ye, and S.~Ji.
\newblock Deep model based transfer and multi-task learning for biological
  image analysis.
\newblock In {\em Proceedings of the 21th ACM SIGKDD International Conference
  on Knowledge Discovery and Data Mining}, pages 1475--1484. ACM, 2015.

\bibitem{zhou2013modeling}
J.~Zhou, J.~Liu, V.~A. Narayan, and J.~Ye.
\newblock {{M}odeling disease progression via multi-task learning}.
\newblock {\em Neuroimage}, 78:233--248, Sep 2013.

\end{thebibliography}
}

\end{document}